\documentclass{article}

\usepackage[preprint]{neurips_2026}

\usepackage[utf8]{inputenc} 
\usepackage[T1]{fontenc}    
\usepackage{hyperref}       
\usepackage{url}            
\usepackage{booktabs}       
\usepackage{amsfonts}       
\usepackage{nicefrac}       
\usepackage{microtype}      
\usepackage{xcolor}         
\usepackage{graphicx}
\usepackage[table]{xcolor} 
\usepackage{enumitem}
\usepackage{subcaption}
\usepackage{multirow}
\usepackage{colortbl}

\usepackage{pifont}

\definecolor{highlightblue}{HTML}{1F77B4} 

\title{Blending Concepts: Benchmarking Visual Metaphor Generation in Text-to-Image Models}

\author{
  Chuer Chen\textsuperscript{1} \quad
  Zichen Wang\textsuperscript{1} \quad
  Yi He\textsuperscript{1} \quad
  Zhengxi Yu\textsuperscript{1} \quad
  Nan Cao\textsuperscript{1,2}\thanks{Corresponding author.}
  \\[4pt]
  \textsuperscript{1}Tongji University
  \qquad
  \textsuperscript{2}Shanghai Innovation Institute
}

\begin{document}

\maketitle

\begin{abstract}
Text-to-image (T2I) models have achieved remarkable success at faithfully rendering specified objects and attributes, yet their ability to produce \emph{visual metaphors}---images that convey abstract ideas by combining elements from two distinct domains---remains largely unexamined. To bridge this gap, we introduce \textbf{VMetaphor-Bench}, the first benchmark for evaluating visual metaphor generation in T2I models. It comprises 1,500 visual metaphors curated from real-world creative imagery, organized into three levels and ten categories, with each sample paired with two prompts of differing specificity. For evaluation, we develop a hybrid framework within an MLLM-as-judge paradigm, combining a multiple-choice question (MCQ) based protocol of 9,594 questions across four levels of metaphorical fidelity with a dimension-based scoring protocol along three perceptual dimensions. Extensive evaluation of 11 representative T2I models reveals that even the strongest proprietary models struggle with compositional structuring and cross-domain mapping, key aspects of metaphorical expression, highlighting visual metaphor generation as an important frontier for future T2I research.
\end{abstract}

\section{Introduction}
Recent years have witnessed remarkable progress in text-to-image (T2I) generation, driven by advances across three main architectural directions: diffusion models~\cite{ho2020denoising,rombach2022high}, autoregressive models~\cite{team2024chameleon,tian2024visual}, and unified multimodal architectures~\cite{chen2025janus, deng2025emerging,wu2025omnigen2}. Powered by large-scale training, a wide range of T2I systems, from leading proprietary models such as GPT-Image-1~\cite{openai2025gptimage} and Gemini 3.1 Flash Image~\cite{google2025gemini} to open-source counterparts~\cite{flux2024,wu2025qwenimage}, now produce photorealistic images that faithfully render specified objects, attributes, and spatial layouts, achieving strong performance on basic semantic prompt following. However, their ability to express abstract concepts and convey meaning beyond what is literally depicted remains far less explored.

\begin{figure}[t]
    \centering
    \includegraphics[width=\textwidth]{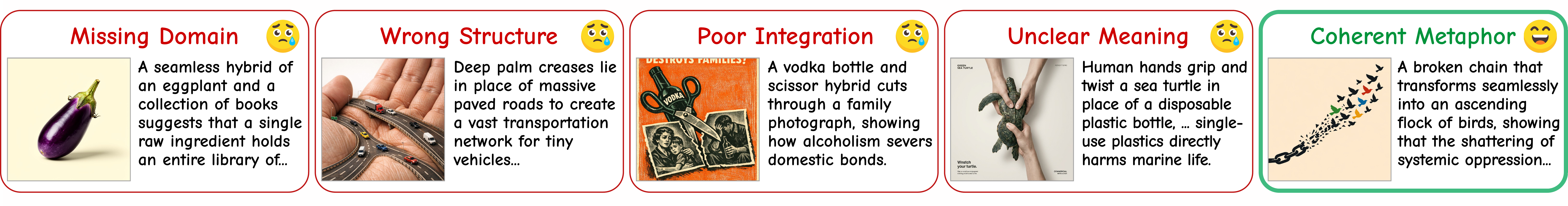}
    \caption{Common failure modes in visual metaphor generation from GPT-Image-1.5~\cite{openai2025gptimage}, Seedream 5.0 Lite~\cite{seedream5lite}, and FLUX.2-dev~\cite{flux-2-2025}.}
    \label{fig:failure}
\end{figure}

A particularly important form of such abstract expression is the visual metaphor, where one concept is understood through another via visual analogy~\cite{forceville2008metaphor}. Drawing on conceptual blending theory~\cite{fauconnier2003conceptual}, a visual metaphor combines elements from two distinct domains — a source that lends its properties, and a target that receives them — to produce a single image carrying meaning richer than either domain alone. Generating such images is far from straightforward: the model must (i) blend the two given domains coherently using a specific compositional strategy (e.g., fusion or juxtaposition), (ii) establish meaningful element-level mappings between the two, and (iii) ensure that the resulting image conveys the intended meaning to viewers. As shown in Figure~\ref{fig:failure}, even state-of-the-art T2I models frequently struggle with these tasks, producing images that miss one of the domains, adopt the wrong structural form, blend the two awkwardly, or fail to convey the intended meaning.

Yet this challenge has been largely overlooked by existing T2I benchmarks, which remain dominated by literal prompt following. Mainstream benchmarks such as GenEval~\cite{ghosh2023geneval} and T2I-CompBench~\cite{huang2023t2i} evaluate object, attribute, and basic spatial fidelity from short prompts, while more recent reasoning-aware benchmarks~\cite{niu2025wise,sun2025t2i,zhao2025envisioning} extend toward factual or physical inference, leaving the cross-domain conceptual blending required by visual metaphors unexamined. On the metaphor side, existing efforts have focused predominantly on the understanding direction~\cite{akula2023metaclue,kundu2025looking,zhang2026metaphorstar}; the only generation evaluation we are aware of, conducted as a small probe within MetaCLUE~\cite{akula2023metaclue}, is limited to 300 samples assessed via FID, CLIP similarity, and a small user study — none of which can systematically diagnose where and how T2I models fail at metaphorical expression. Together, these limitations leave a clear gap: no existing benchmark systematically evaluates how well T2I models generate visual metaphors across diverse themes and metaphorical structures.

\begin{figure}[b]
    \centering
    \includegraphics[width=\textwidth]{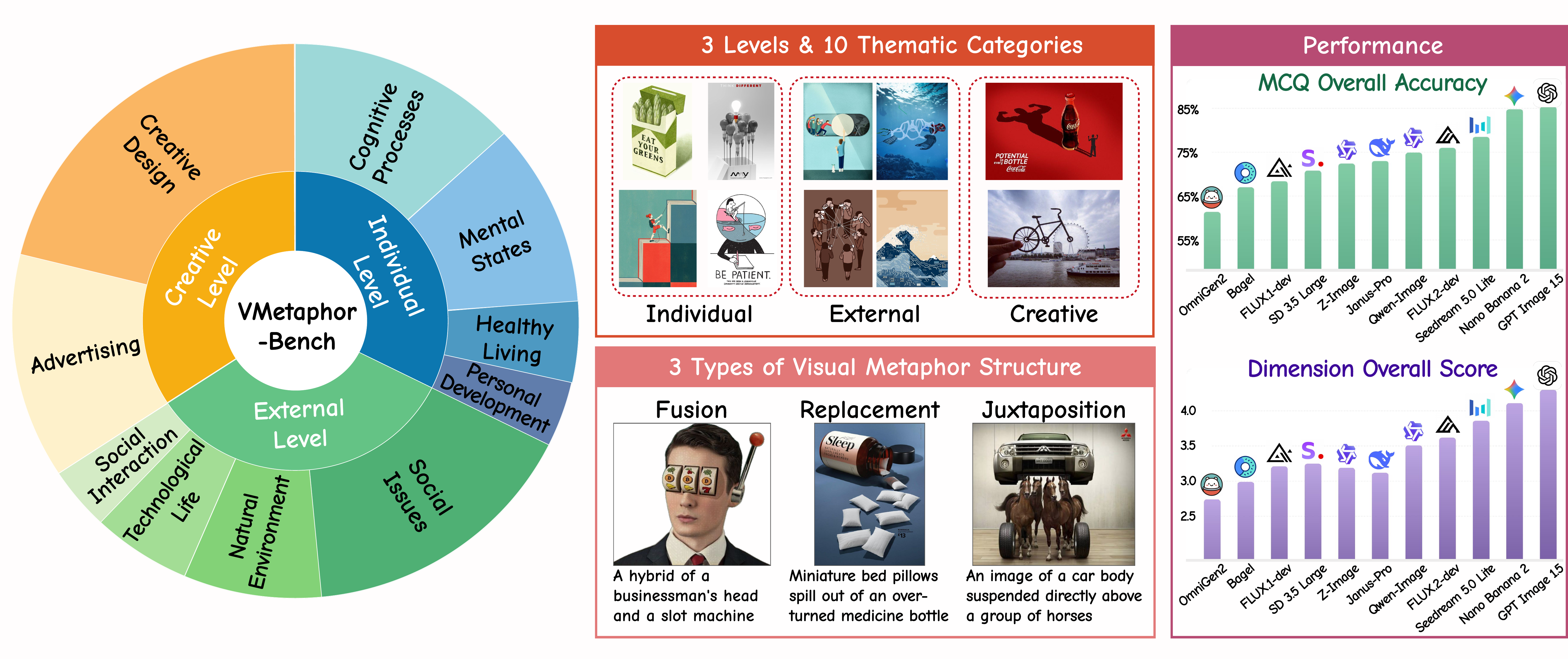}
    \caption{Overview of VMetaphor-Bench. \textbf{Left:} hierarchical taxonomy with three levels and ten thematic categories. \textbf{Middle:} representative examples spanning the three levels and the three structure types of visual metaphors. \textbf{Right:} performance of 11 evaluated T2I models on MCQ accuracy and dimension scoring under conceptual prompts.}
    \label{fig:overview}
\end{figure}

To address this gap, we introduce \textbf{VMetaphor-Bench}, a new benchmark designed for evaluating visual metaphor generation in T2I models. It comprises 1,500 visual metaphors curated from real-world creative imagery, organized into three levels and ten categories, as illustrated in Figure~\ref{fig:overview}. Each sample is annotated with its source and target domains, structure type, metaphorical meaning, element-level mappings, and paired with two prompts of differing specificity: a conceptual prompt that conveys only the metaphorical idea, and a descriptive prompt that details its visual realization. For evaluation, we develop a hybrid framework consisting of an MCQ-based protocol — 9,594 questions targeting four levels of metaphorical fidelity (\textit{domain presence}, \textit{structure type}, \textit{meaning}, and \textit{element mapping}) — and a dimension-based scoring protocol rating each image on \textit{metaphoric efficacy}, \textit{metaphor logic}, and \textit{perceptual harmony}. Both protocols are realized within an MLLM-as-judge paradigm, enabling fine-grained, interpretable diagnosis of T2I models' strengths and weaknesses in visual metaphor generation.

Using VMetaphor-Bench, we conduct a comprehensive evaluation of \textbf{11 representative T2I models}, spanning leading proprietary systems, open-source diffusion models, and unified multimodal architectures. Our results reveal that proprietary models such as GPT Image 1.5 and Nano Banana 2 achieve substantially better performance than open-source counterparts. However, even the strongest models struggle with compositional structuring and cross-domain mapping---key aspects of metaphorical expression---highlighting visual metaphor generation as an important frontier for future T2I research.

In summary, our contributions are threefold:

\begin{itemize}[leftmargin=1.0em]
\item We introduce \textbf{VMetaphor-Bench}, the first benchmark (to the best of our knowledge) dedicated to evaluating visual metaphor generation in T2I models, comprising 1,500 high-quality, carefully curated samples organized into a hierarchical taxonomy of three levels and ten categories.

\item We design a hybrid evaluation framework with an MCQ-based protocol of 9,594 questions across four levels of metaphorical fidelity and a dimension-based protocol along three perceptual dimensions, enabling fine-grained, interpretable assessment.

\item We conduct a comprehensive evaluation of 11 representative T2I models, providing insights into their capabilities in visual metaphor generation and highlighting key directions for future research.
\end{itemize}

\section{Related Works}
\subsection{Text-to-Image Generation Models}
Text-to-image (T2I) generation has advanced rapidly along three main architectural directions. Diffusion models~\cite{dhariwal2021diffusion, ho2020denoising, rombach2022high} remain the dominant paradigm, with recent progress driven by scaling transformer-based denoisers~\cite{esser2024scaling,flux2024,peebles2023scalable}. Autoregressive models~\cite{sun2024autoregressive,team2024chameleon,tian2024visual} cast image generation as sequential token prediction, offering fine-grained compositional control through explicit modeling of inter-token dependencies. More recently, unified multimodal models~\cite{chen2025janus,deng2025emerging, wu2025omnigen2,xie2025show} integrate visual understanding and generation within a single architecture, leveraging multimodal reasoning to better interpret complex prompts. State-of-the-art systems such as GPT-Image-1~\cite{openai2025gptimage} and Gemini 3.1 Flash Image~\cite{google2025gemini} now produce images of remarkable fidelity and aesthetic quality for everyday scenes. However, generating images that convey abstract semantic relationships, such as visual metaphors, remains a significant challenge across all three paradigms.
  
\subsection{Text-to-Image Generation Benchmarks}
Early evaluation of text-to-image generation relied on distribution-level metrics such as FID~\cite{heusel2017gans} and feature-based similarity scores like CLIPScore~\cite{hessel2021clipscore}, which capture perceptual quality but fail to assess fine-grained semantic correctness. The recent rise of multimodal large language models (MLLMs)~\cite{hurst2024gpt,team2023gemini} has enabled a paradigm shift toward MLLM-as-judge evaluation~\cite{chen2024mllm}, where a vision-language model directly examines the generated image against the prompt, providing a more flexible and semantically grounded assessment framework. Under this paradigm, a line of benchmarks targets foundational semantic alignment — verifying whether generated images faithfully depict the specified objects~\cite{kamath2025geneval}, attributes~\cite{ye2025echo}, and spatial relationships~\cite{wang2025genspace, wang2026everything}. Representative works include GenEval~\cite{ghosh2023geneval}, T2I-CompBench~\cite{huang2023t2i}, and TIFA~\cite{hu2023tifa}, which decompose prompts into atomic verification questions to systematically evaluate compositional fidelity. More recently, benchmarks have shifted toward higher-order reasoning capabilities, probing whether T2I models can incorporate world knowledge~\cite{niu2025wise, zhang2025worldgenbench} and reasoning~\cite{chen2025r2i,sun2025t2i, zhao2025envisioning} into their generations, reflecting a broader trend from evaluating object-level fidelity toward scene-level semantic understanding.

\subsection{Visual Metaphor}
Visual metaphor communicates abstract ideas by mapping one conceptual domain onto another through visual imagery~\cite{ervas2021metaphor,forceville2008metaphor,van2014finding}. On the understanding side, MetaCLUE~\cite{akula2023metaclue} introduces a comprehensive benchmark for metaphor classification, localization, interpretation, along with a small probe on generation; ImageMet~\cite{kundu2025looking} benchmarks VLMs on metaphor captioning and VQA, and MetaphorStar~\cite{zhang2026metaphorstar} applies visual reinforcement learning to improve metaphor reasoning. On the generation side, several recent methods leverage text-to-image models as the visual backbone: Creative Blends~\cite{sun2025creative} fuses disparate visual concepts via commonsense knowledge, The Mind's Eye~\cite{koushik2025mind} proposes a source-target-meaning decomposition with self-evaluation rewards, and Beyond Pixels~\cite{xu2026beyond} introduces schema-driven reasoning for metaphor transfer. However, evaluation of these methods has been limited---typically conducted on small-scale ad hoc test sets via FID, CLIP similarity, or user studies, none of which can diagnose where and how T2I models fail at metaphorical expression. While benchmarks exist for metaphor understanding, no counterpart has been established for systematically evaluating the generation quality of visual metaphors.


\section{VMetaphor-Bench}
\begin{figure}[htbp]
    \centering
    \includegraphics[width=\textwidth]{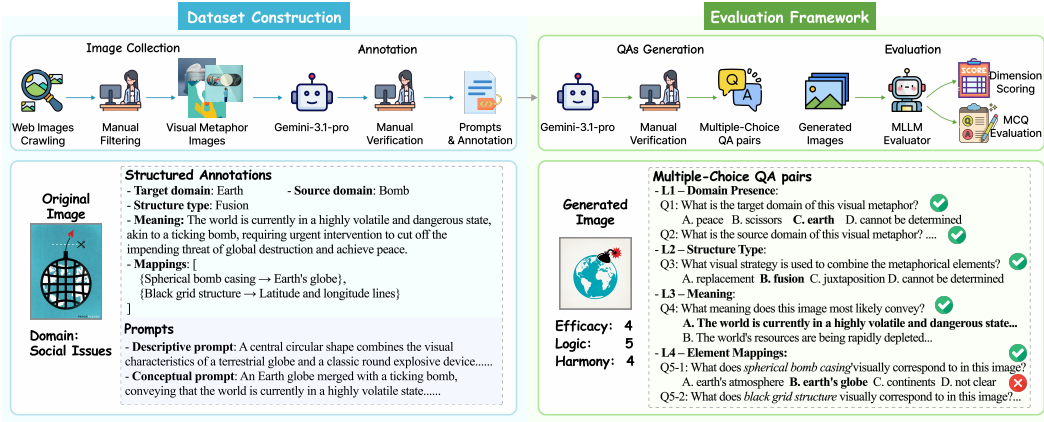}
    \caption{Overview of the VMetaphor-Bench pipeline. \textbf{Left: Dataset Construction} — images are collected and annotated through a two-stage pipeline, producing structured annotations and two granularity levels of prompts. \textbf{Right: Evaluation Framework} — generated images are assessed via MCQ-based evaluation (L1--L4 levels) and dimension-based scoring (Efficacy, Logic, Harmony).}
    \label{fig:pipeline}
\end{figure}

We present VMetaphor-Bench, the first benchmark, to the best of our knowledge, for evaluating visual metaphor generation in text-to-image models, featuring multi-dimensional assessments of cross-domain conceptual blending. We describe the dataset construction process in Section 3.1 and detail the evaluation framework in Section 3.2. An overview of the complete pipeline is illustrated in Figure~\ref{fig:pipeline}.

\subsection{Dataset Construction}
\textbf{Image Collection.} To construct a benchmark grounded in real-world visual metaphors, we collect images from Pinterest~\cite{pinterest}, a platform rich in creative visual content such as advertisements, editorial illustrations, and conceptual art. We query the platform using over ten keywords spanning both general and domain-specific terms, e.g. \textit{visual metaphor}, \textit{creative ads}. This process yields approximately 5,000 candidate images. Two authors then review the entire collection, filtering out duplicates, images that do not contain a clear visual metaphor, and low-quality or ambiguous samples, resulting in approximately 1,600 images for subsequent annotation.

\textbf{Data Annotation.} After collecting images, we employ Gemini 3.1 Pro~\cite{gemini3pro} as the annotation backbone for its strong visual reasoning capabilities. Since visual metaphors involve multi-layered semantics~\cite{cappa2025meanings}, generating accurate annotations and high-quality prompts simultaneously in a single pass is challenging. We therefore adopt a two-stage annotation pipeline. In the first stage, the model is presented with each image and tasked with extracting structured annotations: 
\begin{itemize}[leftmargin=1.5em, itemsep=1pt, topsep=2pt]
  \item \textbf{Target domain}: the subject being described — what the image is ultimately ``about''. 
  \item \textbf{Source domain}: the concept used to characterize the target— it lends a property to the target through visual analogy.
  \item \textbf{Structure type}: how the two domains are visually combined. Following Phillips et al.~\cite{phillips2004beyond}, we categorize it as \textit{replacement} (one domain is absent and inferred from context), \textit{fusion} (both domains are merged into a single object), or \textit{juxtaposition} (both domains appear as separate objects in the same scene).
  \item \textbf{Meaning}: the overall message conveyed through the interaction of the two domains.
  \item \textbf{Element mappings}: fine-grained correspondences specifying which source element maps to which target element.
\end{itemize}

In the second stage, the model receives both the image and the extracted annotations, and generates two prompts at different levels of granularity: (1) The \textit{descriptive prompt} provides a detailed depiction of the visual content and how the metaphor is physically rendered in the image; (2) The \textit{conceptual prompt} specifies the target and source domains, implicitly conveys the structure type, and states the intended meaning, without describing specific visual details or element-level mappings. Additionally, a style suffix describing the visual style of the original image (e.g., photorealistic, illustration) is extracted and appended to both prompts during generation to ensure stylistic consistency. This two-stage design allows the structured annotations to guide prompt generation, ensuring consistency between the structured annotations and the textual descriptions. 

Three annotators, after training on visual metaphor theory and annotation guidelines, perform rigorous manual inspection on all images, annotations, and prompts, and cross-validate each other's decisions. Images where the metaphor is ambiguous or unrecognizable are removed. Inaccurate annotations (e.g., misidentified target or source domains) and imprecise prompts are manually corrected to align with the visual content. This verification process yields our final set of 1,500 high-quality images, each accompanied by structured annotations and two granularity levels of prompts.

\subsection{Evaluation Framework}
Evaluating visual metaphor generation requires assessing both metaphorical fidelity — whether the generated image faithfully captures the intended cross-domain mapping — and expressive quality — how effectively and aesthetically the metaphor is conveyed. To this end, we adopt a hybrid evaluation framework with MLLM-as-judge~\cite{zhang2025large}, comprising two complementary approaches: (1) MCQ-based evaluation that measures the correctness of metaphorical elements through multiple-choice questions, and (2) dimension-based scoring that rates the overall quality of the generated metaphor on a 1-5 Likert scale. An overview of the evaluation framework is shown in Figure~\ref{fig:pipeline} (Right). 

\textbf{MCQ-based Evaluation.} To objectively assess the correctness of generated images at each level of the metaphorical structure, we automatically construct a set of multiple-choice questions (MCQs) for all 1,500 samples in the dataset. Specifically, we leverage Gemini 3.1 Pro to generate MCQs based on each sample's structured annotations along with its descriptive prompt. The generated MCQs assess four hierarchical levels of metaphorical fidelity:
\begin{itemize}[leftmargin=1.5em, itemsep=1pt, topsep=2pt]
    \item \textbf{L1 -- Domain Presence}: Whether the target and source domains are visually recognizable in the image (one question per domain).
    \item \textbf{L2 -- Structure Type}: Whether the visual combination strategy matches the ground-truth structure type (replacement, fusion, or juxtaposition).
    \item \textbf{L3 -- Meaning}: Whether the image conveys the intended metaphorical meaning.
    \item \textbf{L4 -- Element Mapping}: Whether each source element correctly corresponds to its intended target element (one question per mapping pair; scores are averaged).
\end{itemize}

This process yields a total of 9,594 MCQs across all samples, which are manually verified by the same annotators to ensure correctness. Each question consists of four options: one ground-truth answer derived from the annotation, two plausible but incorrect distractors, and a "cannot be determined" option. All option orders are randomized to eliminate positional bias. For each image, we compute per-level accuracy: L1 averages over the two domain questions, L2 is a single binary score, L3 is a single binary score, and L4 averages over all mapping questions. The overall MCQ score is the mean across all four levels.

\textbf{Dimension-based Scoring.} While MCQ-based evaluation measures the correctness of metaphorical elements, it does not capture the overall perceptual quality of the generated image. We therefore complement it with dimension-based scoring, where an MLLM evaluator is presented with the generated image along with its structured annotations and asked to rate the image on a 1--5 Likert scale across three independent dimensions:
  \begin{itemize}[leftmargin=1.5em, itemsep=1pt, topsep=2pt]
    \item \textbf{Metaphoric Efficacy}: How clearly the image communicates the intended metaphorical meaning to a viewer without any textual explanation.
    \item \textbf{Metaphor Logic}: Whether the cross-domain mapping between source and target is conceptually coherent and logically sound.
    \item \textbf{Perceptual Harmony}: How naturally and coherently the two domains are visually presented, considering blending, composition, and overall visual cohesion.
\end{itemize}

The overall dimension score is the average across all three dimensions. Together with the MCQ-based evaluation, this hybrid framework provides a comprehensive assessment of both the metaphorical fidelity and expressive quality of generated visual metaphors. Full evaluation prompts are provided in Appendix~\ref{app:eval_prompts}.
  
\begin{table}[t]
    \caption{Evaluation results on VMetaphor-Bench with \textbf{conceptual prompts}.}
    \label{tab:main_results_conceptual}
    \centering
    \footnotesize
    \setlength{\tabcolsep}{4pt}
    \begin{tabular}{lcccc c ccc c}
      \toprule
      & \multicolumn{5}{c}{MCQ Accuracy (\%)} & \multicolumn{4}{c}{Dimension Score (1--5)} \\
      \cmidrule(lr){2-6} \cmidrule(lr){7-10}
      Model & Domain & Structure & Meaning & Mapping & Overall & Efficacy &
    Logic & Harmony & Overall \\
      \midrule
      \noalign{\vspace{-3pt}}
      \rowcolor{gray!15}
      \multicolumn{10}{l}{\textit{Closed-source T2I Models}} \\
      Seedream 5.0 Lite    & 90.6 & 65.6 & 96.0 & 66.5 & 78.5 & 3.78 & 3.74 & 4.08 & 3.86 \\
      Nano Banana 2        & 93.0 & 76.3 & \textcolor{highlightblue}{\textbf{98.5}} & 75.9 & 84.8 & 4.05 & 4.08 & 4.21 & 4.11 \\
      GPT Image 1.5        & \textcolor{highlightblue}{\textbf{93.6}} & \textcolor{highlightblue}{\textbf{78.1}}  & 97.6 & \textcolor{highlightblue}{\textbf{76.5}} & \textcolor{highlightblue}{\textbf{85.4}} & \textcolor{highlightblue}{\textbf{4.27}} & \textcolor{highlightblue}{\textbf{4.23}} & \textcolor{highlightblue}{\textbf{4.38}} & \textcolor{highlightblue}{\textbf{4.30}} \\
      \midrule
      \noalign{\vspace{-3pt}}
      \rowcolor{gray!15}
      \multicolumn{10}{l}{\textit{Open-source T2I Models}} \\
      FLUX.1-dev           & 82.3 & 54.9 & 85.6 & 55.2 & 68.4 & 2.90 & 2.87 & 3.82 & 3.19 \\
      SD 3.5 Large         & 83.0 & 60.1 & 85.7 & 58.9 & 70.8 & 2.99 & 3.01 & 3.75 & 3.25 \\
      Z-Image              & 84.3 & 59.6 & 89.5 & 60.8 & 72.4 & 2.93 & 3.03 & 3.67 & 3.21 \\
      Qwen-Image           & 87.3 & 66.1 & 90.5 & 61.7 & 74.9 & 3.39 & 3.31 & 3.84 & 3.52 \\ 
      FLUX.2-dev           & 88.6 & 63.7 & 92.1 & 63.9 & 76.0 & 3.48 & 3.46 & 3.93 & 3.62 \\
      \midrule
      \noalign{\vspace{-3pt}}
      \rowcolor{gray!15}
      \multicolumn{10}{l}{\textit{Open-source Unified MLLMs}} \\
      OmniGen2             & 74.8 & 48.1 & 77.1 & 49.1 & 61.4 & 2.53 & 2.44 & 3.45 & 2.74 \\
      Bagel                & 79.8 & 54.8 & 84.2 & 54.2 & 67.0 & 2.69 & 2.72 & 3.56 & 2.99 \\
      Janus-Pro            & 83.4 & 59.1 & 89.1 & 63.2 & 73.0 & 2.90 & 3.02 & 3.43 & 3.12 \\
      \bottomrule
    \end{tabular}
    \vspace{-1em}
\end{table}

\section{Experiments}
\label{sec:experiments}
In this section, we evaluate representative T2I models on VMetaphor-Bench and analyze their behavior under different prompt granularities. We further validate the reliability of our MLLM-as-judge framework through cross-judge consistency and human alignment studies. Additional experimental details and results are provided in the Appendix.

\subsection{Setup}
\textbf{Models.} We evaluate 11 representative text-to-image models spanning three categories: (1) \textbf{Closed-source T2I Models.} GPT Image 1.5~\cite{openai2025gptimage}, Nano Banana 2 (Gemini 3.1 Flash Image)~\cite{google2025gemini}, and Seedream 5.0 Lite~\cite{seedream5lite}. These proprietary models represent the current state of the art in text-to-image generation; (2) \textbf{Open-source T2I Models.} Stable Diffusion 3.5 Large~\cite{sd35large}, FLUX.1-dev~\cite{flux2024}, FLUX.2-dev~\cite{flux-2-2025}, Z-Image~\cite{team2025zimage}, and Qwen-Image~\cite{wu2025qwenimage}. These models represent the leading open-source text-to-image models; (3) \textbf{Open-source Unified MLLMs.} OmniGen2~\cite{wu2025omnigen2}, Bagel~\cite{deng2025emerging}, and Janus-Pro~\cite{chen2025janus}. These models employ a unified architecture for both image generation and multimodal understanding. All models are evaluated using their default generation configurations. 

\textbf{MLLM Evaluator.}  For both MCQ-based evaluation and dimension-based scoring, we adopt an open-source MLLM, Qwen3.5-27B~\cite{qwen3.5}, as the default judge model for its strong visual reasoning capability and full reproducibility. We additionally report results with GPT-5.4~\cite{gpt54} in the Appendix~\ref{app:gpt54} to validate cross-judge consistency.

\subsection{Main Results on VMetaphor-Bench}
We report the main evaluation results with conceptual prompts in Table~\ref{tab:main_results_conceptual}. All metrics are assigned by Qwen3.5-27B as the default judge. Based on the results across 11 models, we highlight several key findings regarding model capabilities and core challenges in visual metaphor generation.

\textbf{Closed-source models lead by a substantial margin.} GPT Image 1.5 achieves the best overall performance with 85.4\% MCQ accuracy and a dimension score of 4.30, followed by Nano Banana 2 (84.8\%, 4.11). In contrast, the best-performing open-source T2I model, FLUX.2-dev, trails by 9.4 percentage points (76.0\%, 3.62). While FLUX.2-dev narrows the gap with the weakest closed-source model Seedream 5.0 Lite (78.5\%), a clear performance tier remains, suggesting that visual metaphor generation—which demands both semantic comprehension and creative compositional ability—still substantially benefits from the larger-scale training and architectural advantages of proprietary systems. Unified MLLMs show mixed performance: while Janus-Pro (73.0\%) is competitive with mid-tier open-source T2I models, OmniGen2 (61.4\%) and Bagel (67.0\%) rank at the bottom despite their multimodal understanding capabilities. 

\textbf{Structure and Mapping are the primary bottleneck.} All models exhibit a consistent difficulty hierarchy across MCQ levels: Domain and Meaning accuracies are consistently the two highest among the four levels for every model, indicating that models can generate images containing the correct conceptual elements and conveying the intended metaphorical meaning. However, performance drops sharply on Structure (48.1--78.1\%) and Mapping (49.1--76.5\%), revealing that the core challenge lies in compositional structuring—models can depict the right elements but struggle to compose them into coherent metaphorical structures. This pattern is particularly evident in Seedream 5.0 Lite, which achieves 90.6\% on Domain and 96.0\% on Meaning yet drops to 65.6\% on Structure and 66.5\% on Mapping. Notably, a deeper analysis of structure type errors reveals that \emph{replacement} is far more challenging than \emph{fusion} or \emph{juxtaposition}, a pattern that mirrors the cognitive complexity hierarchy in visual rhetoric theory~\cite{phillips2004beyond} (see Appendix~\ref{app:per_structure}).

\textbf{Visually harmonious but metaphorically shallow.} Across all models, Perceptual Harmony is consistently the highest-scoring dimension (3.43--4.38), while Metaphoric Efficacy and Metaphor Logic generally trail behind. This suggests that although the intended meaning can be inferred (as reflected by high meaning accuracy), the metaphor is not conveyed with sufficient clarity and immediacy through visual structure—models produce aesthetically pleasing images without powerful metaphorical expression. The gap between Harmony and Efficacy widens for weaker models: OmniGen2 scores 3.45 on Harmony but only 2.53 on Efficacy (gap = 0.92), whereas GPT Image 1.5 maintains a gap of only 0.11 (4.38 vs.\ 4.27).

\begin{figure}[t]
    \centering
    \includegraphics[width=\textwidth]{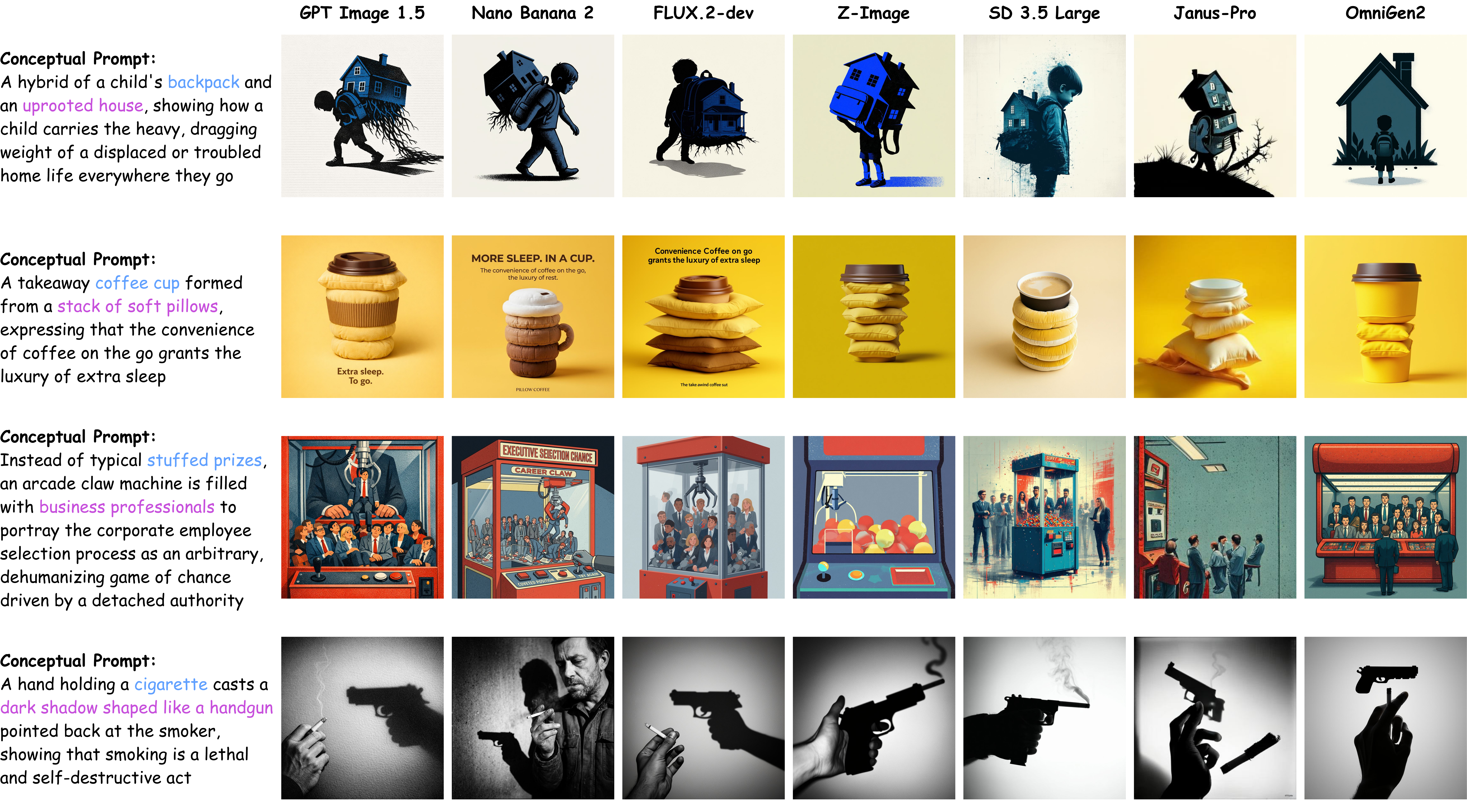}
    \caption{Qualitative comparison of generated images across seven representative models on four visual metaphors. Each row corresponds to one metaphor, with the conceptual prompt shown on the left. The four metaphors cover three structure types: fusion (rows 1--2), replacement (row 3), and juxtaposition (row 4). Blue and purple text highlight the target and source domains, respectively.}
    \label{fig:qualitative}
    \vspace{-1em}
\end{figure}

\subsection{Qualitative Analysis}
To complement the quantitative results above, Figure~\ref{fig:qualitative} presents representative outputs from seven models on four visual metaphors covering different structure types (fusion, replacement, and juxtaposition). Several patterns emerge that are consistent with our quantitative findings. Closed-source models such as GPT Image 1.5 and Nano Banana 2 generally produce coherent compositions that integrate the two domains in meaningful ways. In contrast, weaker open-source models often struggle to compose the 
two domains correctly. Mid-tier models such as FLUX.2-dev and Z-Image fall in between, occasionally capturing the metaphorical concept but producing visually disharmonious compositions or failing on structurally demanding cases.

A closer look at typical failure cases reveals three recurring issues that map directly to our quantitative findings. The most common is \emph{incorrect structural composition}: models depict both domains but fail to combine them in the structural form implied by the prompt. For example, in row 4, where the prompt suggests a juxtaposition of a cigarette and a handgun-shaped shadow, SD 3.5 Large, Z-Image, and Janus-Pro instead fuse the two into a single physical object---echoing the sharp performance drop on L2 Structure across all models. A second issue is \emph{poor visual integration}, where the two domains are placed together but blend awkwardly, such as a coffee cup stacked on visibly distinct pillows without any unified form (row 2). Although the resulting images can still appear visually polished, the awkward integration weakens the metaphor's expressive force, which is consistent with our finding that Metaphoric Efficacy lags behind Perceptual Harmony across most models. In the most severe cases, models \emph{omit one of the domains} entirely: Z-Image, for instance, fills the claw machine with ordinary toy balls rather than business professionals (row 3), making the metaphor impossible to recover. Together, these failure modes underscore the value of a multi-level evaluation framework for diagnosing the specific weaknesses of current T2I models.

\subsection{Effect of Prompt Granularity}
In addition to the conceptual prompts reported in our main experiments, we also generate and evaluate images using the descriptive prompts paired with each sample. Full results under descriptive prompts are provided in the Appendix~\ref{app:descriptive}. Here we analyze the differences between the two prompt settings on a representative subset of five models spanning the three model categories, summarized in Table~\ref{tab:prompt_comparison}.

We first observe that all models benefit from descriptive prompts on MCQ accuracy, and most models also see improvements on dimension scores, indicating that explicit visual specification reduces the difficulty of metaphor generation. However, the magnitude of improvement varies dramatically across models. GPT Image 1.5 gains only +3.1 percentage points on MCQ overall and shows essentially no change on dimension score ($-$0.06), whereas Z-Image improves by +16.6 points and +0.69, respectively. A clear inverse relationship emerges between a model's baseline capability and the gain it receives: stronger models exhibit smaller gains, while weaker models benefit substantially more.

This pattern suggests that capable models such as GPT Image 1.5 and Nano Banana 2 already possess the ability to reason about \emph{how} a metaphorical idea should be visually composed without explicit specification. In contrast, weaker models lack this inferential capacity and rely heavily on the explicit visual guidance provided by descriptive prompts. This aligns with our earlier observation that visual metaphor generation requires more than literal rendering: it demands a degree of conceptual reasoning, on which current open-source models still fall short.

Beyond the per-model gains, descriptive prompts also narrow the capability gap between models. Under conceptual prompts, the dimension score gap between GPT Image 1.5 (4.30) and Z-Image (3.21) is 1.09; under descriptive prompts, this gap narrows to 0.34 (4.24 vs.\ 3.90). For practitioners working with open-source models, providing fine-grained visual descriptions is therefore an effective strategy to compensate for the limited metaphorical reasoning capability.

\begin{table}[t]
    \caption{Performance comparison between \textbf{conceptual} and \textbf{descriptive} prompts on representative models. ``Conc.'' and ``Desc.'' denote the conceptual and descriptive prompt settings, respectively. $\Delta$ denotes the change when switching from conceptual to descriptive prompts.}
    \label{tab:prompt_comparison}
    \centering
    \footnotesize
    \begin{tabular}{lccc c ccc}
      \toprule
      & \multicolumn{3}{c}{MCQ Overall (\%)} & & \multicolumn{3}{c}{Dimension Overall} \\
      \cmidrule(lr){2-4} \cmidrule(lr){6-8}
      Model & Conc. & Desc. & $\Delta$ & & Conc. & Desc. & $\Delta$ \\
      \midrule
      GPT Image 1.5  & 85.4 & 88.5 & +3.1  & & 4.30 & 4.24 & $-$0.06 \\
      Nano Banana 2  & 84.8 & 89.2 & +4.4  & & 4.11 & 4.18 & +0.07 \\
      FLUX.2-dev     & 76.0 & 87.3 & +11.3 & & 3.62 & 4.09 & +0.47 \\
      Z-Image        & 72.4 & 85.9 & +13.5 & & 3.21 & 3.90 & +0.69 \\
      OmniGen2       & 61.4 & 78.0 & +16.6 & & 2.74 & 3.35 & +0.61 \\
      \bottomrule
    \end{tabular}
    \vspace{-2.0em}
\end{table}

\subsection{Reliability of the Evaluator}
To validate the reliability of using Qwen3.5-27B as our default evaluator, we conduct two studies: a cross-judge consistency study assessing robustness to the choice of evaluator, and a human alignment study measuring agreement with human judgment.

\textbf{Cross-Judge Consistency.} 
To test whether our conclusions depend on the specific choice of evaluator, we re-run the full evaluation pipeline with GPT-5.4~\cite{gpt54} as an alternative judge and compare model rankings and scores against those produced by Qwen3.5-27B. As shown in Table~\ref{tab:cross_judge_short}, the top-5 models receive identical rankings from both judges across both metrics. Over all 11 models, Spearman's $\rho$ reaches 0.991 for MCQ and 0.936 for dimension scores, confirming that our findings are not dependent on the choice of evaluator.

\begin{table}[t]
  \centering
  \caption{Cross-judge comparison. Model rankings by MCQ Overall and Dimension
  Overall under two judges.}
  \label{tab:cross_judge_short}
  \scriptsize
  \begin{tabular}{lcccc}
  \toprule
  & \multicolumn{2}{c}{MCQ} & \multicolumn{2}{c}{Dimension} \\
  \cmidrule(lr){2-3} \cmidrule(lr){4-5}
  Model & Qwen3.5 & GPT-5.4 & Qwen3.5 & GPT-5.4 \\
  \midrule
  GPT Image 1.5   & 1 & 1 & 1 & 1 \\
  Nano Banana 2   & 2 & 2 & 2 & 2 \\
  Seedream 5.0 Lite & 3  & 3  & 3  & 3  \\
  FLUX.2-dev      & 4  & 4  & 4  & 4  \\
  Qwen-Image      & 5  & 5  & 5  & 5  \\
  \bottomrule
  \end{tabular}
\end{table}

\textbf{Human Alignment Study.} 
We further evaluate how well the two MLLM judges align with human annotators. We randomly sample 100 images each from FLUX.2-dev and Qwen-Image, yielding 200 images in total. Five human annotators, trained on visual metaphor theory and our annotation guidelines, independently evaluate each image using the same protocol applied to the MLLM evaluator. We measure alignment using balanced accuracy~\cite{brodersen2010balanced} for MCQ evaluation and Mean Absolute Error (MAE) for dimension-based scoring, following established practice in recent benchmark studies~\cite{wang2026everything, zhao2025envisioning}. Results are reported in Table~\ref{tab:human_alignment}. 

The MCQ alignment is consistently high across all four levels, with both Qwen3.5-27B and GPT-5.4 achieving over 80\% balanced accuracy overall. Alignment is highest on the more concrete L1 (Domain) and L3 (Meaning) levels (above 83\%), and lower on L2 (Structure) and L4 (Mapping), which require finer-grained judgments about how source and target are visually combined. For dimension-based scoring, Qwen3.5-27B yields a low MAE across all three dimensions (below 0.8 on a 1--5 scale), indicating that its ratings are close to those of human annotators. GPT-5.4 shows slightly larger errors (up to 0.89 on Logic) but remains within a reasonable range (< 1.0). The fact that an open-source evaluator achieves alignment comparable to or better than a leading proprietary model further supports the choice of Qwen3.5-27B as our default judge.

\begin{table}[t]
    \centering
    \scriptsize 
    \caption{Alignment between MLLM evaluators (Qwen3.5-27B and GPT-5.4) and human annotators on 200 sampled images. (a) MCQ alignment by balanced accuracy (\%); (b) dimension alignment by mean absolute error (MAE) on a 1--5 scale.}
    \label{tab:human_alignment}
    
    \begin{subtable}[t]{0.5\linewidth}
      \centering
      \caption{MCQ alignment (Bal.Acc, \%)}
      \label{tab:human_alignment_mcq}
      \begin{tabular}{lccccc}
        \toprule
        Judge & Domain & Structure & Meaning & Mapping & Overall \\
        \midrule
        Qwen3.5-27B & 87.61 & 71.25 & 83.97 & 76.45 & 80.24 \\
        GPT-5.4     & 88.79 & 73.89 & 84.50 & 76.51 & 81.16 \\
        \bottomrule
      \end{tabular}
    \end{subtable}%
    \hfill
    \begin{subtable}[t]{0.40\linewidth}
      \centering
      \caption{Dimension alignment (MAE)}
      \label{tab:human_alignment_dim}
      \begin{tabular}{lccc}
        \toprule
        Judge & Efficacy & Logic & Harmony \\
        \midrule
        Qwen3.5-27B & 0.65 & 0.71 & 0.64 \\
        GPT-5.4     & 0.71 & 0.89 & 0.75 \\
        \bottomrule
      \end{tabular}
    \end{subtable}
\end{table}

\section{Conclusion}
In this paper, we introduce VMetaphor-Bench, the first 
benchmark for evaluating visual metaphor generation in text-to-image models. Built upon 1,500 carefully curated visual metaphors with structured annotations, our benchmark provides a hierarchical taxonomy of three levels and ten categories and a hybrid evaluation framework that combines MCQ-based fidelity assessment with dimension-based scoring. Extensive evaluation of 11 representative T2I models reveals that proprietary models substantially outperform open-source counterparts, yet even the strongest models continue to struggle with compositional structuring and cross-domain mapping---key aspects of metaphorical expression. We hope VMetaphor-Bench serves as a foundation for future research on bridging the gap between literal 
rendering and creative, meaning-rich visual generation.



\bibliographystyle{plain}
\bibliography{references}
  







\appendix

\section{Appendix}
\subsection{Benchmark Statistics}
\label{app:benchmark_statistics}
Table~\ref{tab:dataset_stats} summarizes the dataset statistics. Conceptual prompts are concise (mean 25.6 words), while descriptive prompts provide richer detail (mean 65.0 words); appending the style suffix roughly doubles the length in both cases. We organize the 1,500 images into 10 thematic categories across three levels: \textit{Individual} (495 images) covers topics related to personal cognition, emotion, health, and growth; \textit{External} (499 images) encompasses societal, environmental, technological, and interpersonal themes; and \textit{Creative} (506 images) includes artistic designs and commercial advertisements. The distribution is approximately balanced across levels, with the largest single category being Creative Design (21.07\%).

\begin{table}[htbp]
  \centering
  \caption{Left: Prompt length statistics (word count). Right: Distribution of thematic categories.}
  \scriptsize 
  \label{tab:dataset_stats}
  \begin{minipage}[t]{0.45\linewidth}
  \centering
  \begin{tabular}{lccc}
  \toprule
   & Mean & Min & Max \\
  \midrule
  Conceptual Prompt & 25.6 & 10 & 56 \\
  \quad + Style Suffix & 68.4 & 29 & 133 \\
  Descriptive Prompt & 65.0 & 20 & 142 \\
  \quad + Style Suffix & 107.7 & 38 & 226 \\
  \bottomrule
  \end{tabular}
  \end{minipage}
  \hfill
  \begin{minipage}[t]{0.50\linewidth}
  \centering
  \begin{tabular}{llrc}
  \toprule
  Level & Category & Count & Prop.(\%) \\
  \midrule
  \multirow{4}{*}{\shortstack[l]{Individual\\(495)}}
   & Cognitive Processes & 198 & 13.20 \\
   & Mental States & 161 & 10.73 \\
   & Healthy Living & 72 & 4.80 \\
   & Personal Development & 64 & 4.27 \\
  \midrule
  \multirow{4}{*}{\shortstack[l]{External\\(499)}}
   & Social Issues & 243 & 16.20 \\
   & Natural Environment & 119 & 7.93 \\
   & Technological Life & 88 & 5.87 \\
   & Social Interaction & 49 & 3.27 \\
  \midrule
  \multirow{2}{*}{\shortstack[l]{Creative\\(506)}}
   & Creative Design & 316 & 21.07 \\
   & Advertising & 190 & 12.67 \\
  \bottomrule
  \end{tabular}
  \end{minipage}
\end{table}

\subsection{Experimental Setup Details}
\label{app:model_details}
We conduct all experiments on a single server equipped with 4 $\times$ NVIDIA H200 GPUs. Detailed information about each evaluated model and MLLM evaluators is provided in Table~\ref{tab:model_details}.

\begin{table}[t]
    \centering
    \footnotesize
    \caption{Detailed information about the evaluated T2I models and MLLM evaluators.}
    \label{tab:model_details}
    \setlength{\tabcolsep}{4pt}
    \begin{tabular}{llccl}
      \toprule
      Model & Year & Resolution & Source & Version / HF Checkpoint \\
      \midrule
      \rowcolor{gray!15}
      \multicolumn{5}{l}{\textit{Closed-source T2I Models}} \\
      Seedream 5.0 Lite & 2026 & ~2K & API & \texttt{doubao-seedream-5-0-260128} \\
      Nano Banana 2 & 2026 & 1024 $\times$ 1024 & API & \texttt{gemini-3.1-flash-image-preview} \\
      GPT Image 1.5 & 2025 & 1024 $\times$ 1024 & API & \texttt{gpt-image-1.5-2025-12-16} \\
      \midrule
      \rowcolor{gray!15}
      \multicolumn{5}{l}{\textit{Open-source T2I Models}} \\
      FLUX.1-dev & 2024 & 1024 $\times$ 1024 & Checkpoint & \texttt{black-forest-labs/FLUX.1-dev} \\
      SD 3.5 Large & 2024 & 1024 $\times$ 1024 & Checkpoint & \texttt{stabilityai/stable-diffusion-3.5-large} \\
      Z-Image & 2025 & 1024 $\times$ 1024 & Checkpoint & \texttt{Tongyi-MAI/Z-Image-Turbo} \\
      Qwen-Image & 2025 & 1328 $\times$ 1328 & Checkpoint & \texttt{Qwen/Qwen-Image-2512} \\
      FLUX.2-dev & 2025 & 1024 $\times$ 1024 & Checkpoint & \texttt{black-forest-labs/FLUX.2-dev} \\
      \midrule
      \rowcolor{gray!15}
      \multicolumn{5}{l}{\textit{Open-source Unified MLLMs}} \\
      OmniGen2 & 2025 & 1024 $\times$ 1024 & Checkpoint & \texttt{OmniGen2/OmniGen2} \\
      Bagel & 2025 & 1024 $\times$ 1024 & Checkpoint & \texttt{ByteDance-Seed/BAGEL-7B-MoT} \\
      Janus-Pro & 2025 & 384 $\times$ 384 & Checkpoint & \texttt{deepseek-ai/Janus-Pro-7B} \\
      \midrule
      \rowcolor{gray!15}
      \multicolumn{5}{l}{\textit{MLLM Evaluators}} \\
      Qwen3.5-27B & 2026 & -- & Checkpoint & \texttt{Qwen/Qwen3.5-27B} \\
      GPT-5.4 & 2026 & -- & API & \texttt{gpt-5.4-2026-03-05} \\
      \bottomrule
    \end{tabular}
\end{table}

\subsection{Per-Structure Analysis}
\label{app:per_structure}

\begin{figure}[htbp]
    \centering
    \includegraphics[width=0.6\textwidth]{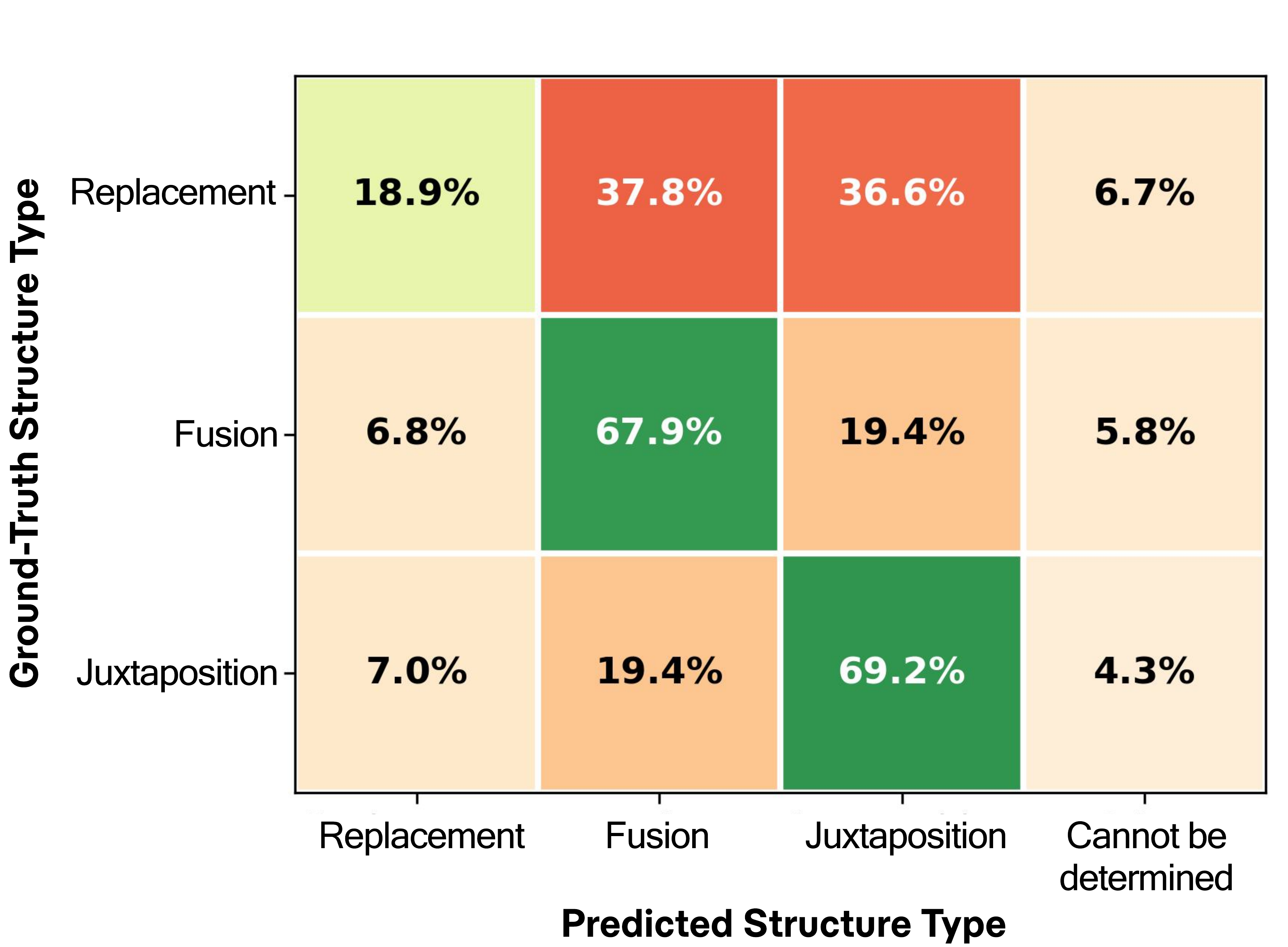}
    \caption{Confusion matrix of predicted vs. ground-truth structure types, aggregated across all models.}
    \label{fig:confusion_matrix}
\end{figure}

To better understand where models struggle on the L2 Structure level, we examine the structure type predicted by the MCQ evaluation against the ground-truth structure type annotated in our dataset. Each L2 MCQ question asks the judge to identify the structure type of the generated image among \emph{fusion}, \emph{replacement}, \emph{juxtaposition}, and \emph{cannot be determined}, allowing us to construct a confusion matrix between ground-truth and predicted structure types. Aggregated results across all 11 evaluated models are shown in Figure~\ref{fig:confusion_matrix}.

The confusion matrix reveals a striking asymmetry across structure types. Models achieve reasonable accuracy on \emph{fusion} (67.9\%) and \emph{juxtaposition} (69.2\%), where the two domains are either merged into a single object or placed alongside each other. In contrast, \emph{replacement} accuracy drops dramatically to only 18.9\%. When prompted with replacement metaphors, models tend to either fuse the two domains together (37.8\%) or place them as separate objects (36.6\%), rather than substituting one for the other.

Interestingly, this difficulty hierarchy mirrors the cognitive complexity ordering proposed in visual rhetoric theory~\cite{phillips2004beyond}, where replacement is identified as the most demanding structure type for human viewers because the missing element must be inferred from its absence.  In contrast, fusion requires disentangling two co-present elements, while juxtaposition simply pairs them side by side. Our results extend this ordering from comprehension to generation: T2I models exhibit the same complexity gradient when producing visual metaphors. This suggests that the cognitive demands of replacement, which require preserving context while substituting a localized element, pose a fundamental challenge for current models. This finding points to compositional precision, the ability to perform localized, structure-aware modifications, as a key direction for advancing visual metaphor generation.

\subsection{Full Results with Descriptive Prompts}
\label{app:descriptive}

Table~\ref{tab:main_results_descriptive} reports the full evaluation results under descriptive prompts, complementing the conceptual prompt results in Table~\ref{tab:main_results_conceptual}. Most models show 
substantial improvements under descriptive prompts, as the detailed visual descriptions provide more explicit guidance for generation. A notable exception is GPT Image 1.5, whose dimension score remains essentially unchanged ($-$0.06), suggesting that the strongest proprietary models can already infer appropriate visual realizations from abstract conceptual prompts and do not require further specification. Despite these magnitude differences, the overall model ranking remains largely consistent across the two prompt types.

\begin{table}[htbp]
      \caption{Evaluation results on VMetaphor-Bench with \textbf{descriptive prompts}.}
      \label{tab:main_results_descriptive}
      \centering
      \footnotesize
      \setlength{\tabcolsep}{4pt}
      \begin{tabular}{lcccc c ccc c}
        \toprule
        & \multicolumn{5}{c}{MCQ Accuracy (\%)} & \multicolumn{4}{c}{Dimension Score (1--5)} \\
        \cmidrule(lr){2-6} \cmidrule(lr){7-10}
        Model & Domain & Structure & Meaning & Mapping & Overall & Efficacy & Logic & Harmony & Overall \\
        \midrule
        \noalign{\vspace{-3pt}}
        \rowcolor{gray!15}
        \multicolumn{10}{l}{\textit{Closed-source T2I Models}} \\
        Seedream 5.0 Lite    & 89.3 & 69.2 & 96.5 & 87.4 & 86.6 & 4.02 & 4.14 & 4.27 & 4.15 \\
        GPT Image 1.5        & 89.6 & 72.9 & 96.6 & 90.6 & 88.5 & \textcolor{highlightblue}{\textbf{4.12}} & \textcolor{highlightblue}{\textbf{4.24}} & \textcolor{highlightblue}{\textbf{4.34}} & \textcolor{highlightblue}{\textbf{4.24}} \\
        Nano Banana 2        & \textcolor{highlightblue}{\textbf{90.8}} & \textcolor{highlightblue}{\textbf{73.6}} & \textcolor{highlightblue}{\textbf{97.1}} & \textcolor{highlightblue}{\textbf{91.2}} & \textcolor{highlightblue}{\textbf{89.2}} & 4.06 & 4.21 & 4.28 & 4.18 \\
        \midrule
        \noalign{\vspace{-3pt}}
        \rowcolor{gray!15}
        \multicolumn{10}{l}{\textit{Open-source T2I Models}} \\
        SD 3.5 Large         & 83.0 & 62.3 & 93.4 & 76.0 & 78.8 & 3.36 & 3.45 & 3.82 & 3.54 \\
        FLUX.1-dev           & 83.8 & 64.1 & 93.5 & 79.6 & 80.7 & 3.45 & 3.55 & 3.97 & 3.66 \\
        Z-Image              & 87.9 & 67.7 & 96.4 & 87.5 & 85.9 & 3.76 & 3.90 & 4.04 & 3.90 \\
        Qwen-Image           & 88.6 & 70.9 & 95.9 & 88.5 & 87.0 & 3.98 & 4.03 & 4.08 & 4.03 \\
        FLUX.2-dev           & 89.4 & 70.3 & 96.4 & 88.9 & 87.3 & 4.00 & 4.09 & 4.18 & 4.09 \\
        \midrule
        \noalign{\vspace{-3pt}}
        \rowcolor{gray!15}
        \multicolumn{10}{l}{\textit{Open-source Unified MLLMs}} \\
        OmniGen2             & 82.6 & 59.9 & 93.2 & 75.3 & 78.0 & 3.07 & 3.26 & 3.71 & 3.35 \\
        Bagel                & 83.2 & 62.7 & 92.6 & 76.8 & 79.1 & 3.28 & 3.36 & 3.77 & 3.47 \\
        Janus-Pro            & 84.0 & 65.9 & 94.9 & 79.1 & 81.0 & 3.31 & 3.52 & 3.62 & 3.48 \\
        \bottomrule
      \end{tabular}
\end{table}

\subsection{Cross-Judge Results with GPT-5.4}
\label{app:gpt54}
Table~\ref{tab:main_results_gpt54} reports the full evaluation results using GPT-5.4 as the judge, complementing the Qwen3.5-27B results in Table~\ref{tab:main_results_conceptual}. Across all 11 models, GPT-5.4 produces highly consistent rankings with Qwen3.5-27B, with Spearman rank correlations of 0.991 for MCQ and 0.936 for dimension scoring. The two judges differ slightly in absolute scores, with GPT-5.4 tending to assign somewhat more lenient dimension scores. However, the relative performance ordering of models is preserved, confirming that the conclusions reported in our main experiments do not depend on the specific choice of evaluator.

\begin{table}[htbp]
  \caption{Evaluation results on VMetaphor-Bench with \textbf{conceptual prompts}, judged by \textbf{GPT-5.4}.}
  \label{tab:main_results_gpt54}
  \centering
  \footnotesize
  \setlength{\tabcolsep}{4pt}
  \begin{tabular}{lcccc c ccc c}
  \toprule
  & \multicolumn{5}{c}{MCQ Accuracy (\%)} & \multicolumn{4}{c}{Dimension
Score (1--5)} \\
   \cmidrule(lr){2-6} \cmidrule(lr){7-10}
  Model & Domain & Structure & Meaning & Mapping & Overall & Efficacy & Logic & Harmony & Overall \\
  \midrule
  \noalign{\vspace{-3pt}}
  \rowcolor{gray!15}
  \multicolumn{10}{l}{\textit{Closed-source T2I Models}} \\ 
   Seedream 5.0 Lite    & 93.3 & 66.7 & 96.3 & 73.0 & 82.0 & 3.72 & 3.85 & 3.94 & 3.84 \\
   Nano Banana 2        & 94.3 & 75.6 & \textcolor{highlightblue}{\textbf{98.3}} & 76.9 & 85.5 & 4.08 & 4.29 & 4.14 & 4.17 \\
   GPT Image 1.5        & \textcolor{highlightblue}{\textbf{94.4}} & \textcolor{highlightblue}{\textbf{77.2}} & 97.6 & \textcolor{highlightblue}{\textbf{77.2}} & \textcolor{highlightblue}{\textbf{85.8}} & \textcolor{highlightblue}{\textbf{4.13}} & \textcolor{highlightblue}{\textbf{4.37}} & \textcolor{highlightblue}{\textbf{4.33}} & \textcolor{highlightblue}{\textbf{4.27}} \\
    \midrule
    \noalign{\vspace{-3pt}}
    \rowcolor{gray!15}
    \multicolumn{10}{l}{\textit{Open-source T2I Models}} \\
    FLUX.1-dev           & 91.3 & 63.8 & 91.9 & 62.8 & 76.4 & 2.95 & 3.07 & 3.70 & 3.24 \\
    SD 3.5 Large         & 91.5 & 65.7 & 91.4 & 64.6 & 77.4 & 2.97 & 3.14 & 3.68 & 3.26 \\
    Z-Image              & 93.1 & 66.5 & 94.8 & 65.6 & 78.9 & 3.05 & 3.25 & 3.56 & 3.28 \\
    Qwen-Image           & 91.9 & 70.3 & 93.9 & 67.3 & 79.6 & 3.32 & 3.45 & 3.73 & 3.50 \\
    FLUX.2-dev           & 94.0 & 69.3 & 94.7 & 69.0 & 80.9 & 3.45 & 3.62 & 3.84 & 3.64 \\
    \midrule
    \noalign{\vspace{-3pt}}
    \rowcolor{gray!15}
    \multicolumn{10}{l}{\textit{Open-source Unified MLLMs}} \\
    OmniGen2             & 90.2 & 62.0 & 90.9 & 57.6 & 73.7 & 2.51 & 2.65 & 3.42 & 2.86 \\
    Bagel                & 91.7 & 64.7 & 92.6 & 62.1 & 76.6 & 2.78 & 2.92 & 3.44 & 3.05 \\
    Janus-Pro            & 92.3 & 67.1 & 93.9 & 67.6 & 79.4 & 3.06 & 3.34 & 3.49 & 3.30 \\
    \bottomrule
  \end{tabular}
\end{table}

\subsection{Benchmark Construction Prompts.}
\begin{figure}[htbp]
    \centering
    \includegraphics[width=0.85\textwidth]{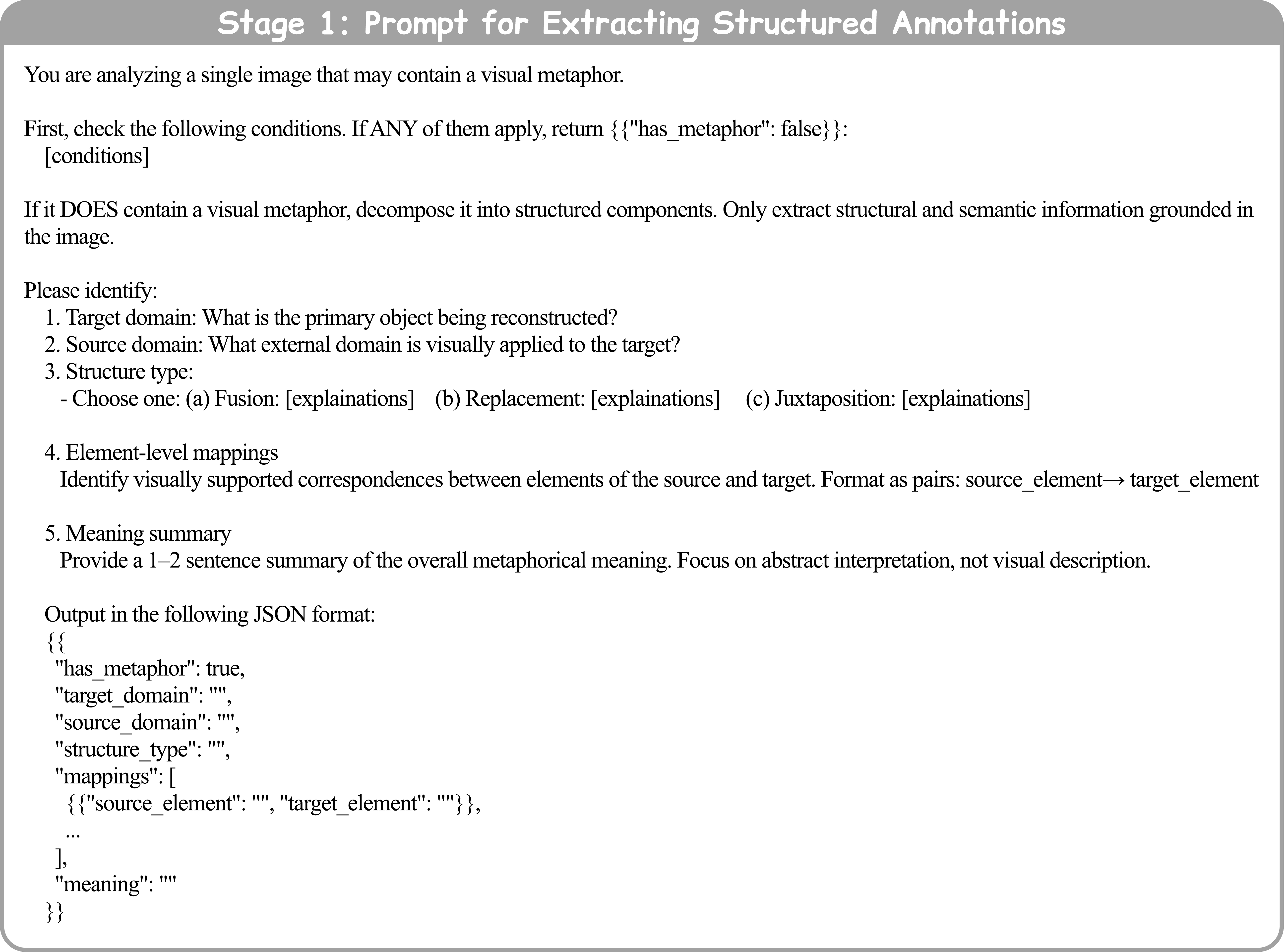}
    \caption{Prompt for structured annotation generation (Stage 1).}
    \label{fig:prompt_annotate_structured}
\end{figure}

\begin{figure}[htbp]
    \centering
    \includegraphics[width=0.85\textwidth]{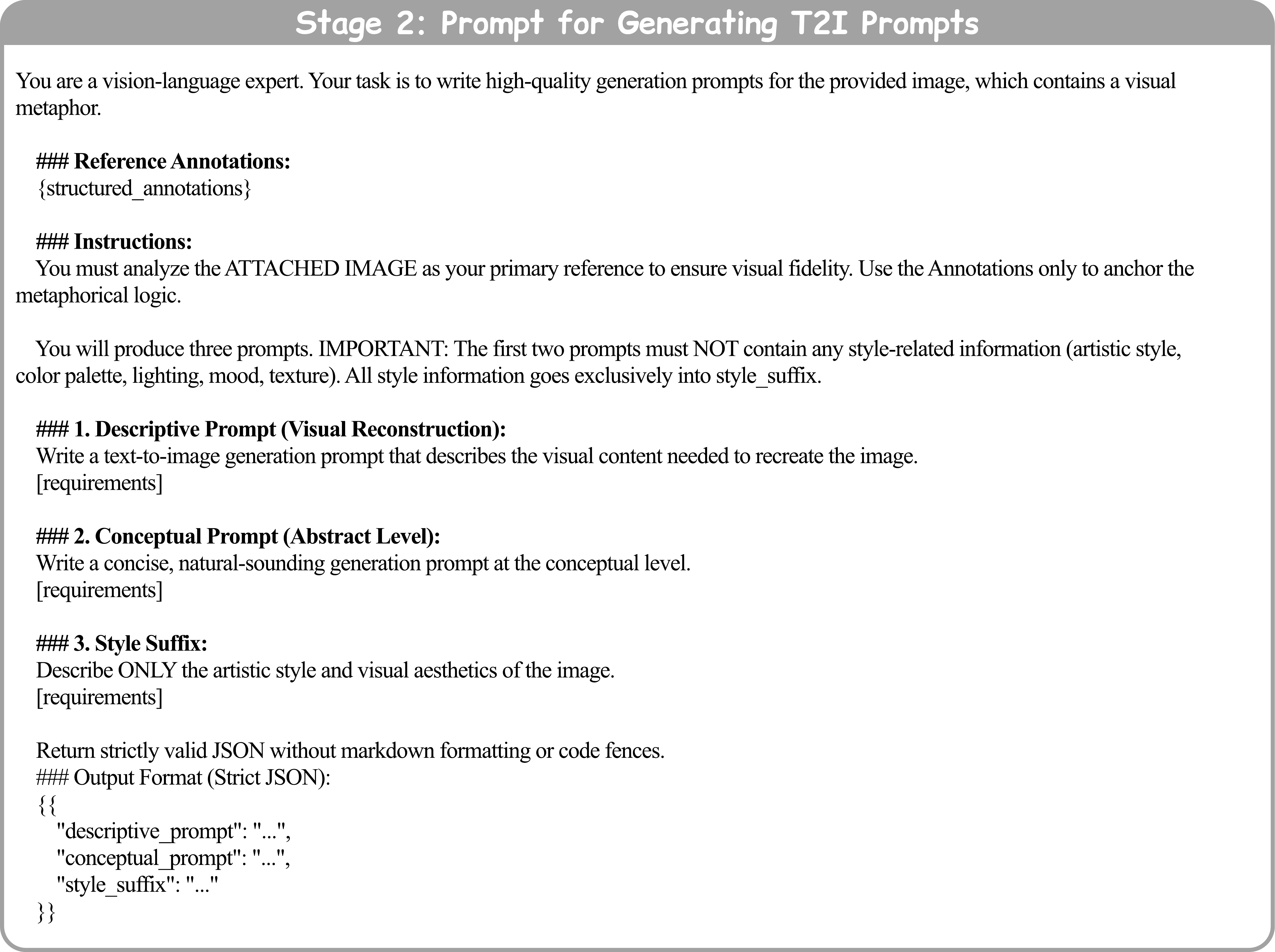}
    \caption{Prompt for descriptive and conceptual prompt generation (Stage 2).}
    \label{fig:prompt_annotate_prompt}
\end{figure}

\begin{figure}[htbp]
    \centering
    \includegraphics[width=0.85\textwidth]{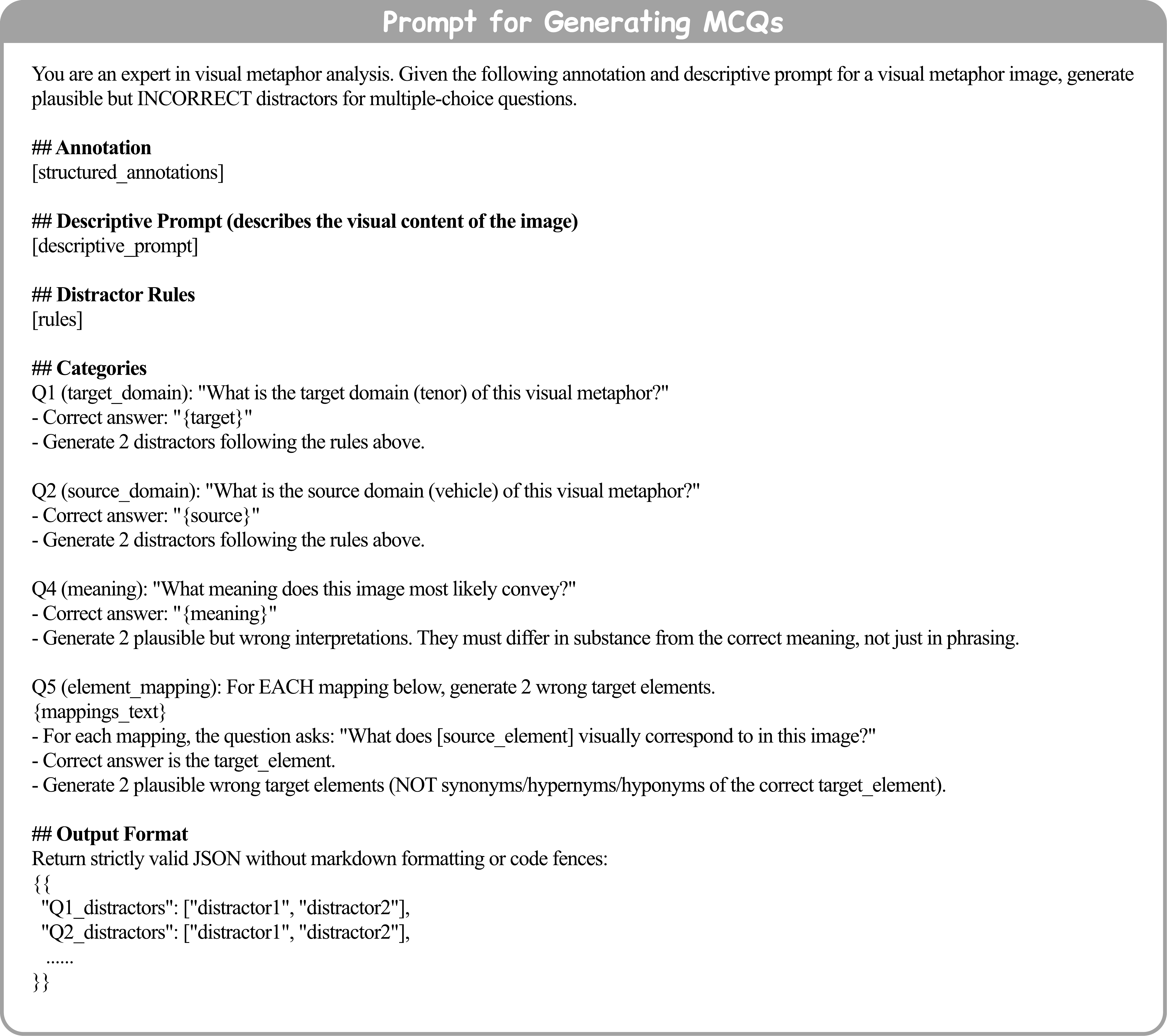}
    \caption{Prompt for MCQ distractor generation.}
    \label{fig:prompt_generate_mcq}
\end{figure}

Figure~\ref{fig:prompt_annotate_structured} and Figure~\ref{fig:prompt_annotate_prompt} show the prompts for the two-stage
annotation pipeline: structured annotation extraction (Stage 1) and descriptive/conceptual prompt generation (Stage 2). Figure~\ref{fig:prompt_generate_mcq} shows the prompt for MCQ distractor generation.
   
\clearpage

\clearpage
\subsection{Evaluation Prompts}
\label{app:eval_prompts}
\begin{figure}[htbp]
    \centering
    \includegraphics[width=0.85\textwidth]{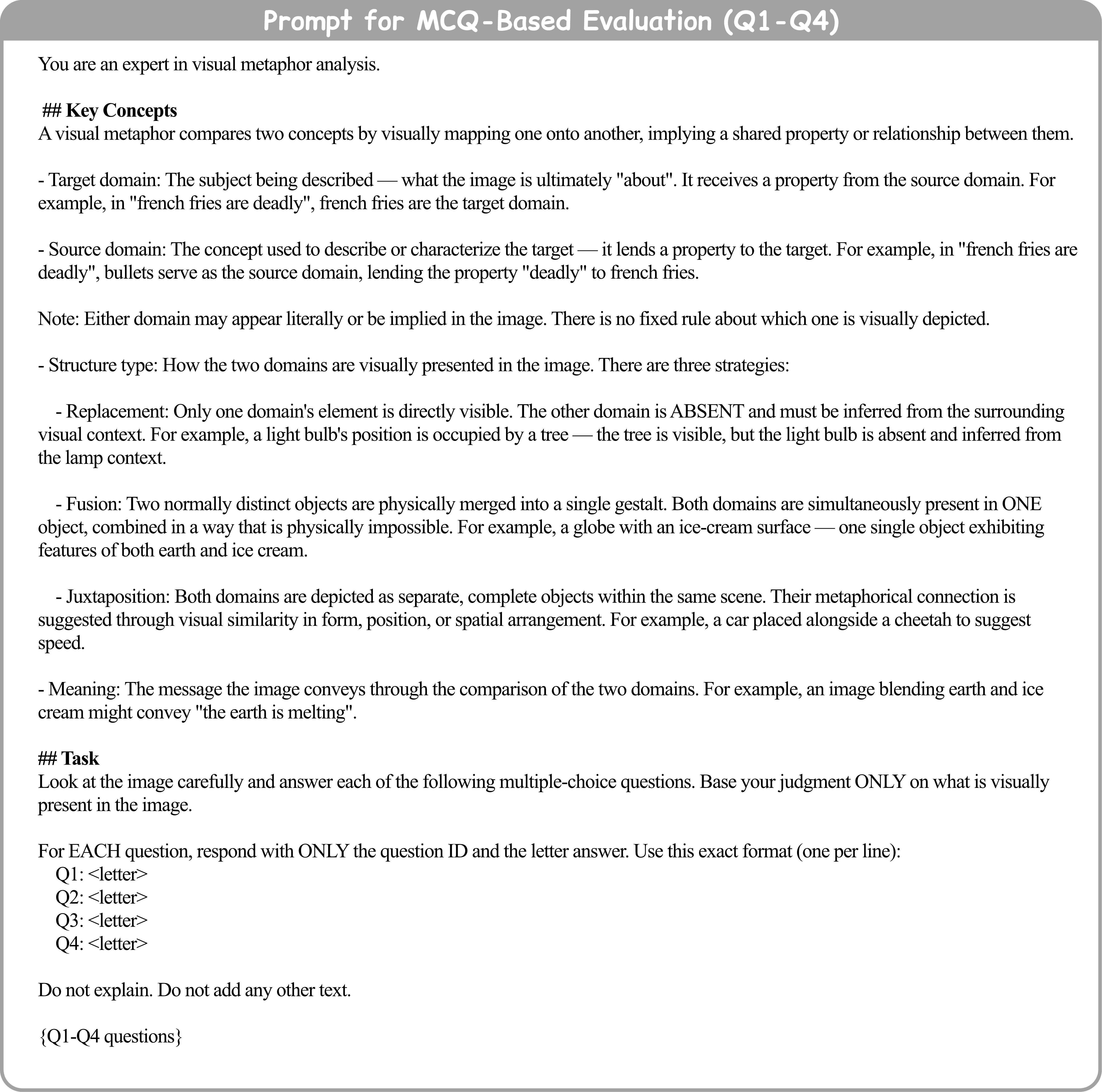}
    \caption{Prompt for MCQ evaluation (Q1--Q4: target domain, source domain, structure type, and meaning).}
    \label{fig:prompt_mcq}
\end{figure}

\begin{figure}[htbp]
    \centering
    \includegraphics[width=0.85\textwidth]{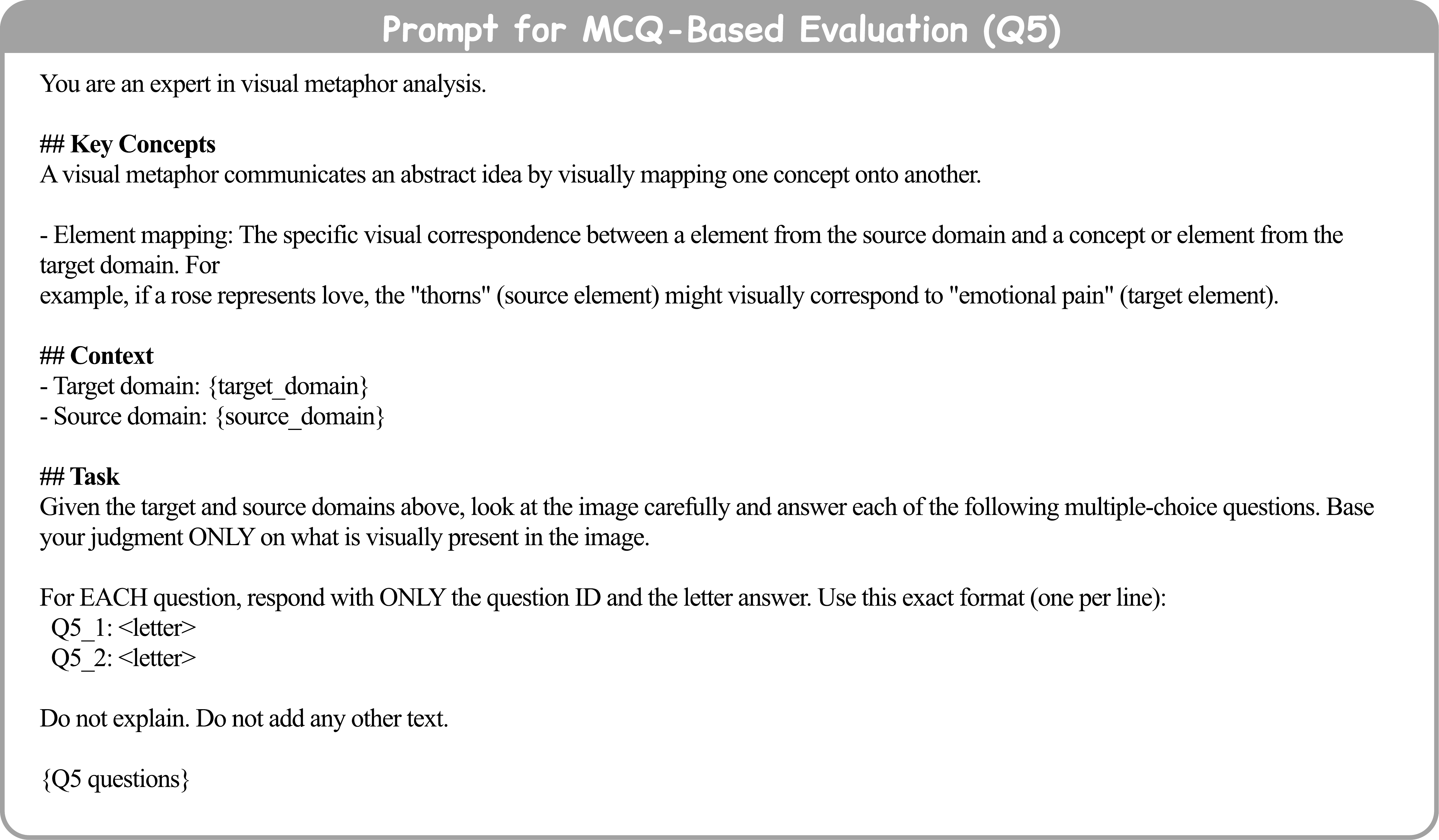}
    \caption{Prompt for MCQ evaluation (Q5: Element mapping).}
    \label{fig:prompt_mcq_q5}
\end{figure}

\begin{figure}[htbp]
    \centering
    \includegraphics[width=0.85\textwidth]{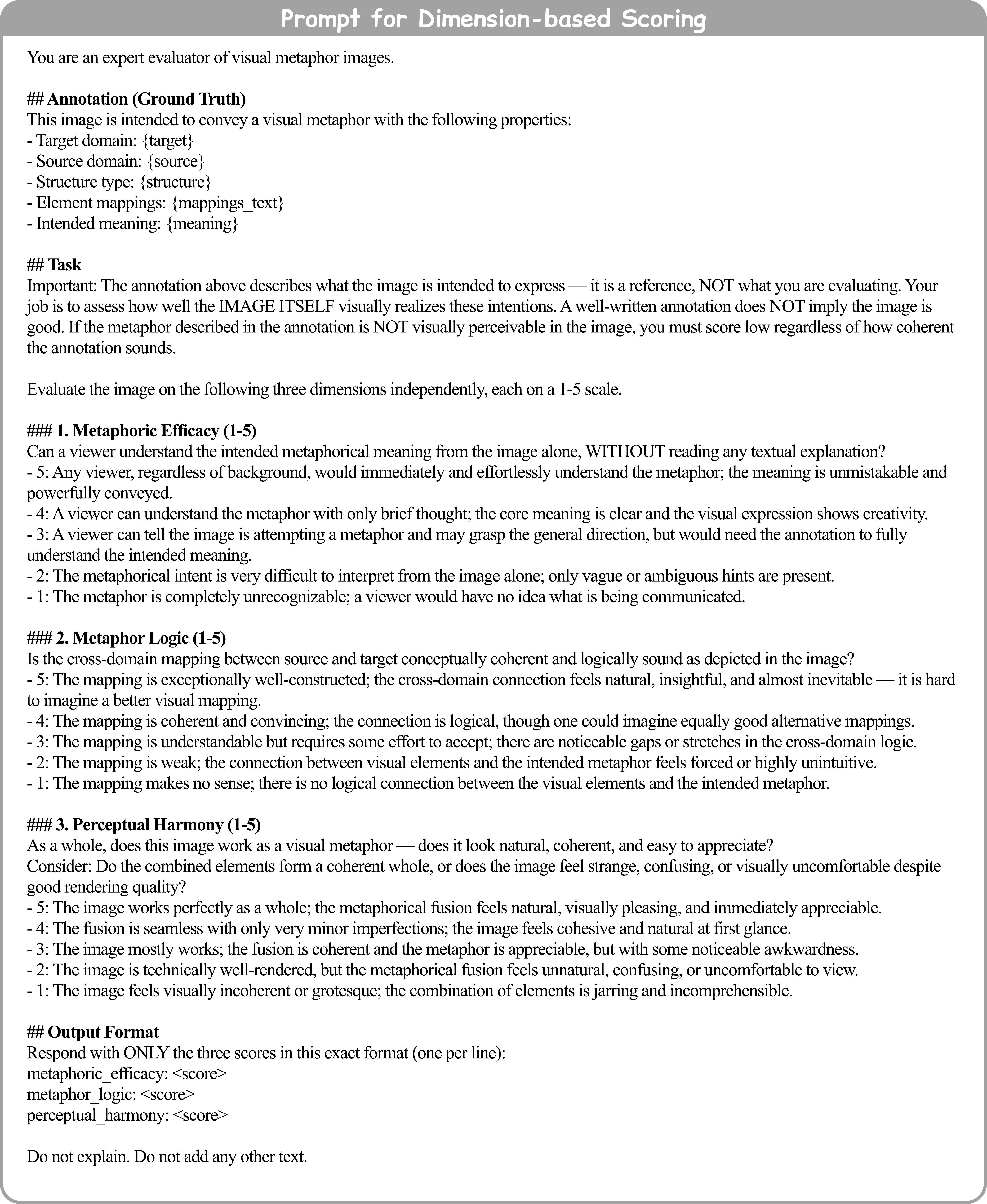}
    \caption{Prompt for dimension score evaluation.}
    \label{fig:prompt_dimension}
\end{figure}

Figure~\ref{fig:prompt_mcq} and Figure~\ref{fig:prompt_mcq_q5} show the prompts for MCQ evaluation, which is split into two calls: Q1--Q4 (target, source, structure, and meaning) are answered together, while Q5 (element mapping) is answered separately to prevent domain leakage from its answer options. Specifically, Q5 (element mapping) options may reveal the source domain (e.g., a question stem mentioning 'spherical bomb casing' would expose 'bomb' as the source), which would compromise the independence of L1 and L4 evaluations. Figure~\ref{fig:prompt_dimension} shows the prompt for dimension scoring with 1--5 Likert-scale rubrics.

\clearpage
\subsection{Interface for Human Annotators}
\label{app:interface}
\begin{figure}[htbp]
    \centering
    \includegraphics[width=\textwidth]{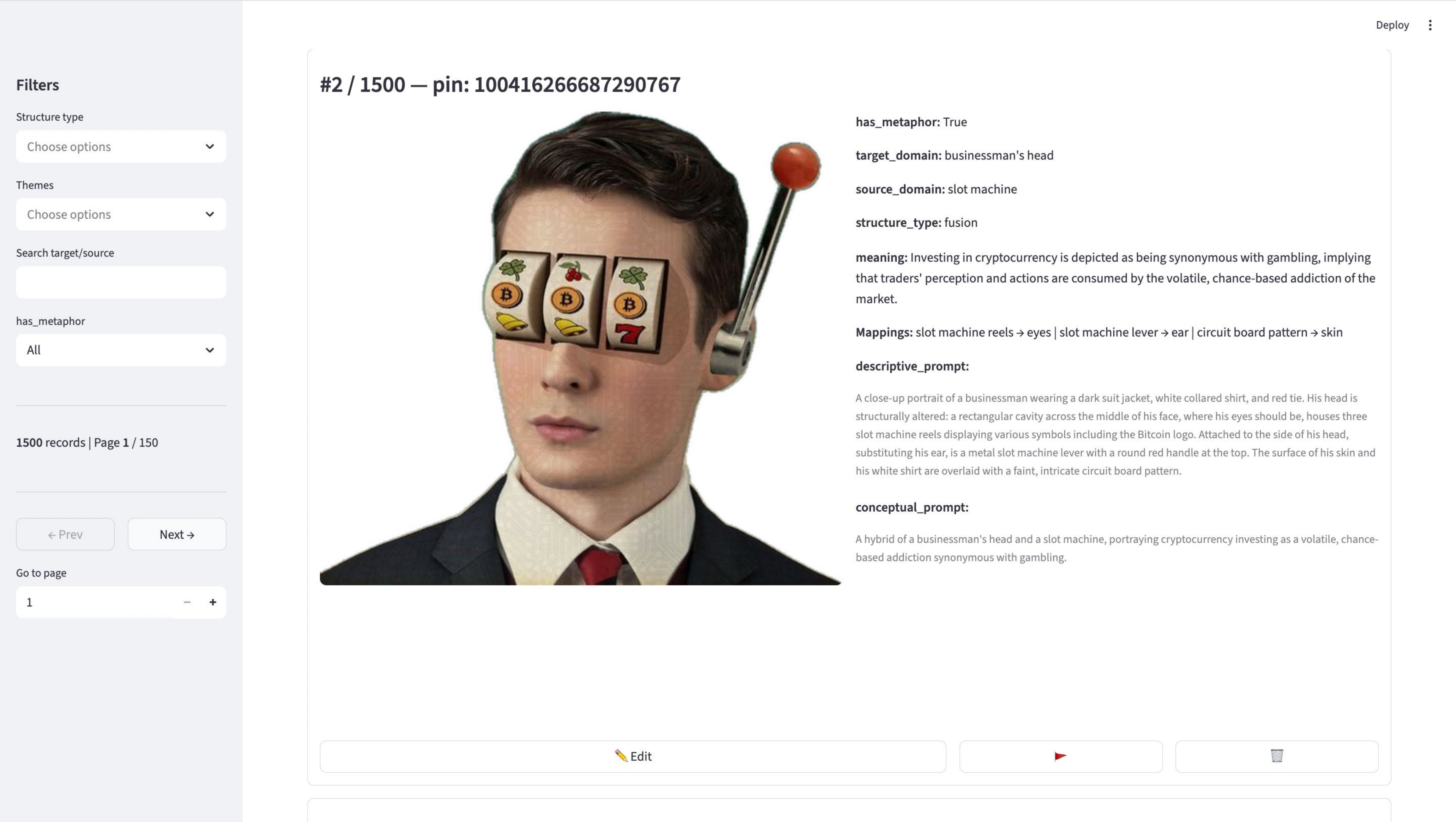}
    \caption{Manual verification interface for data curation}
    \label{fig:annotation_interface}
\end{figure}

\begin{figure}[htbp]
    \centering
    \includegraphics[width=\textwidth]{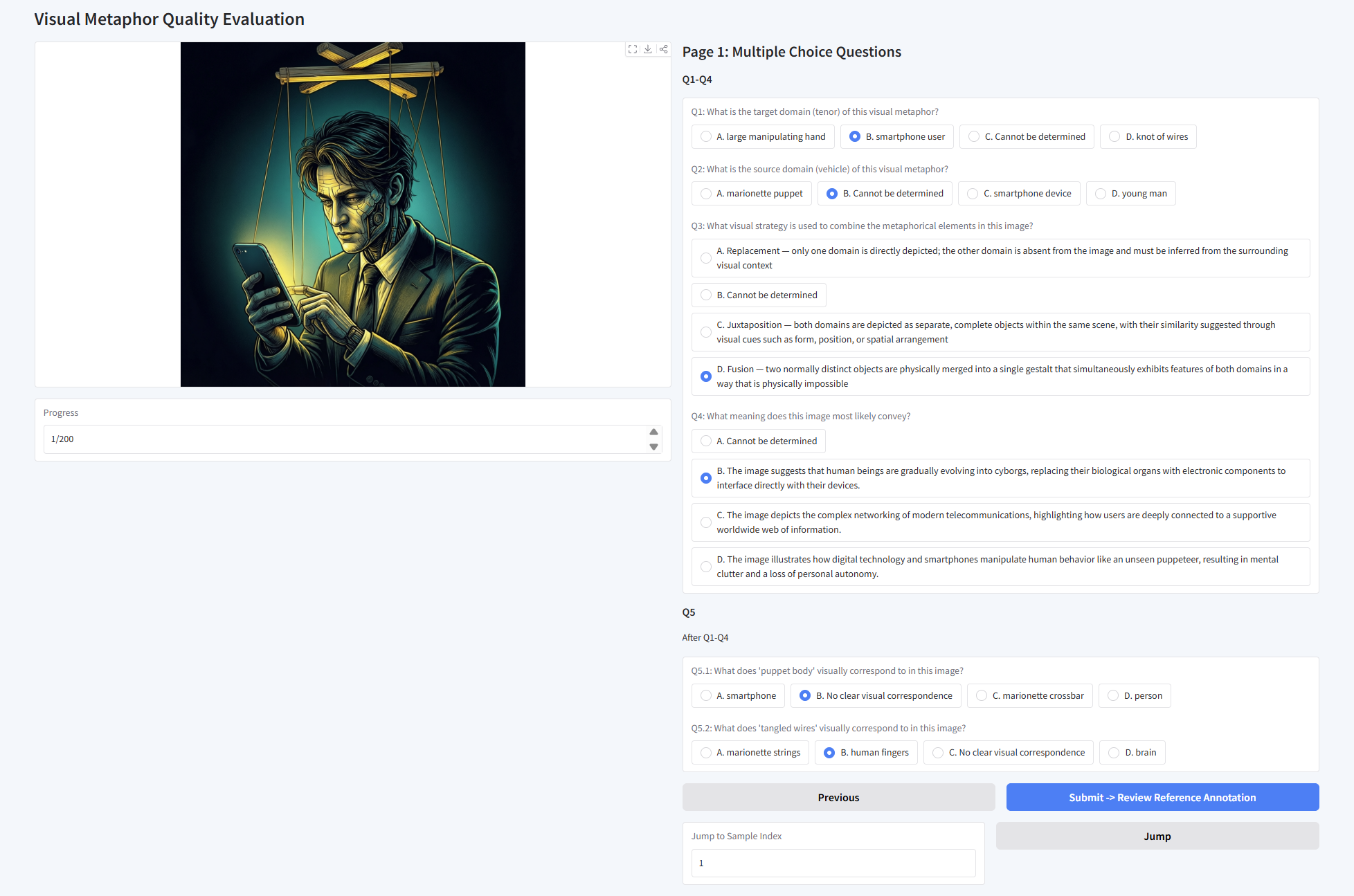}
    \caption{Human evaluation interface for the alignment study: MCQ answering}
    \label{fig:human_align_interface1}
\end{figure}

\begin{figure}[htbp]
    \centering
    \includegraphics[width=\textwidth]{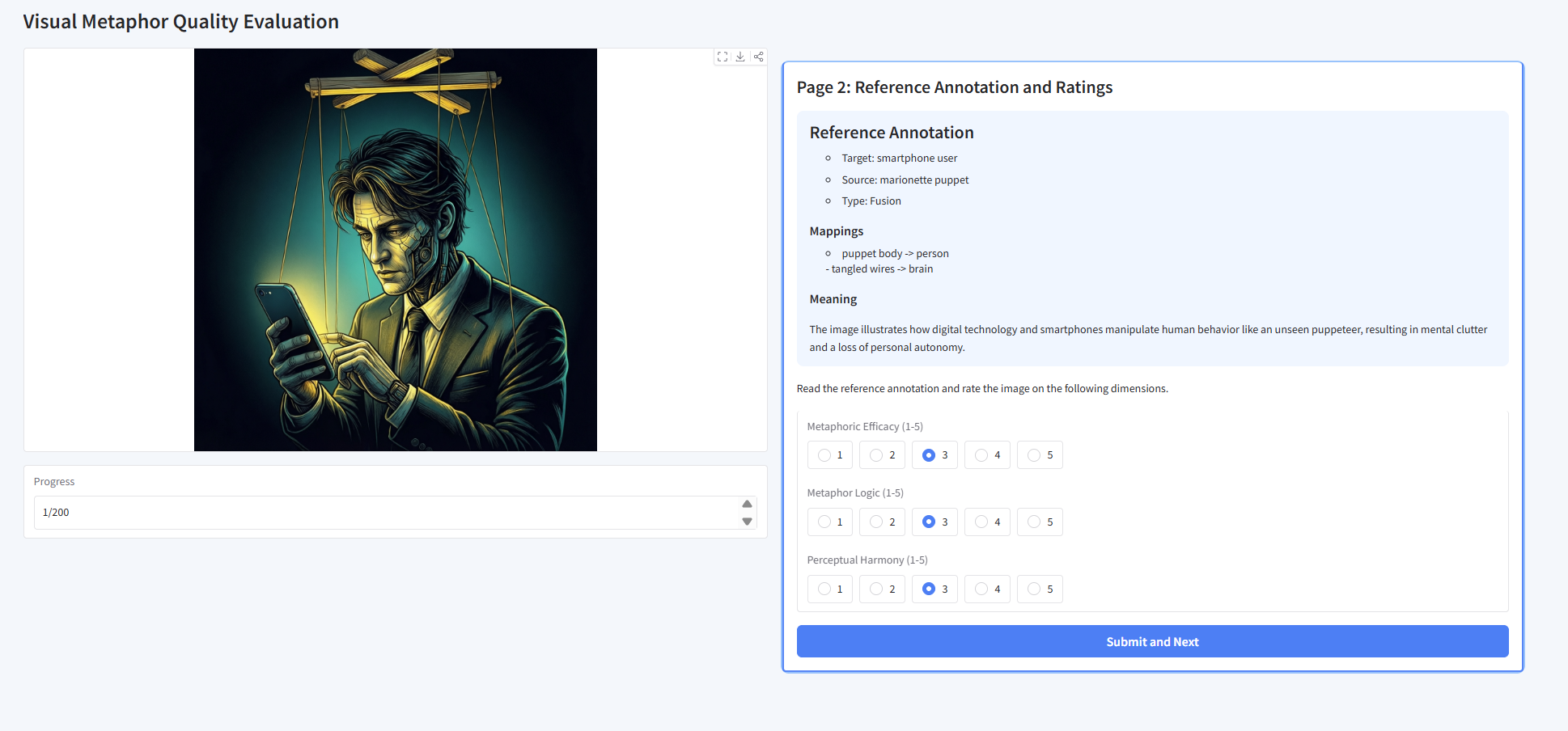}
    \caption{Human evaluation interface for the alignment study: dimension rating}
    \label{fig:human_align_interface2}
\end{figure}

We provide screenshots of the annotation interfaces used in our study. Figure~\ref{fig:annotation_interface} shows the interface for manual verification during data curation (Section 3.1), where annotators review and edit metadata fields such as domains, structure type, element mappings, and meaning. Figures~\ref{fig:human_align_interface1} and \ref{fig:human_align_interface2} show the two pages of the interface used in the human alignment study (Section 4.5): one for answering MCQ questions alongside the generated image, and another for rating dimension scores on a 1--5 scale.
  
\clearpage
\subsection{Samples}
\begin{figure}[htbp]
    \centering
    \includegraphics[width=\textwidth]{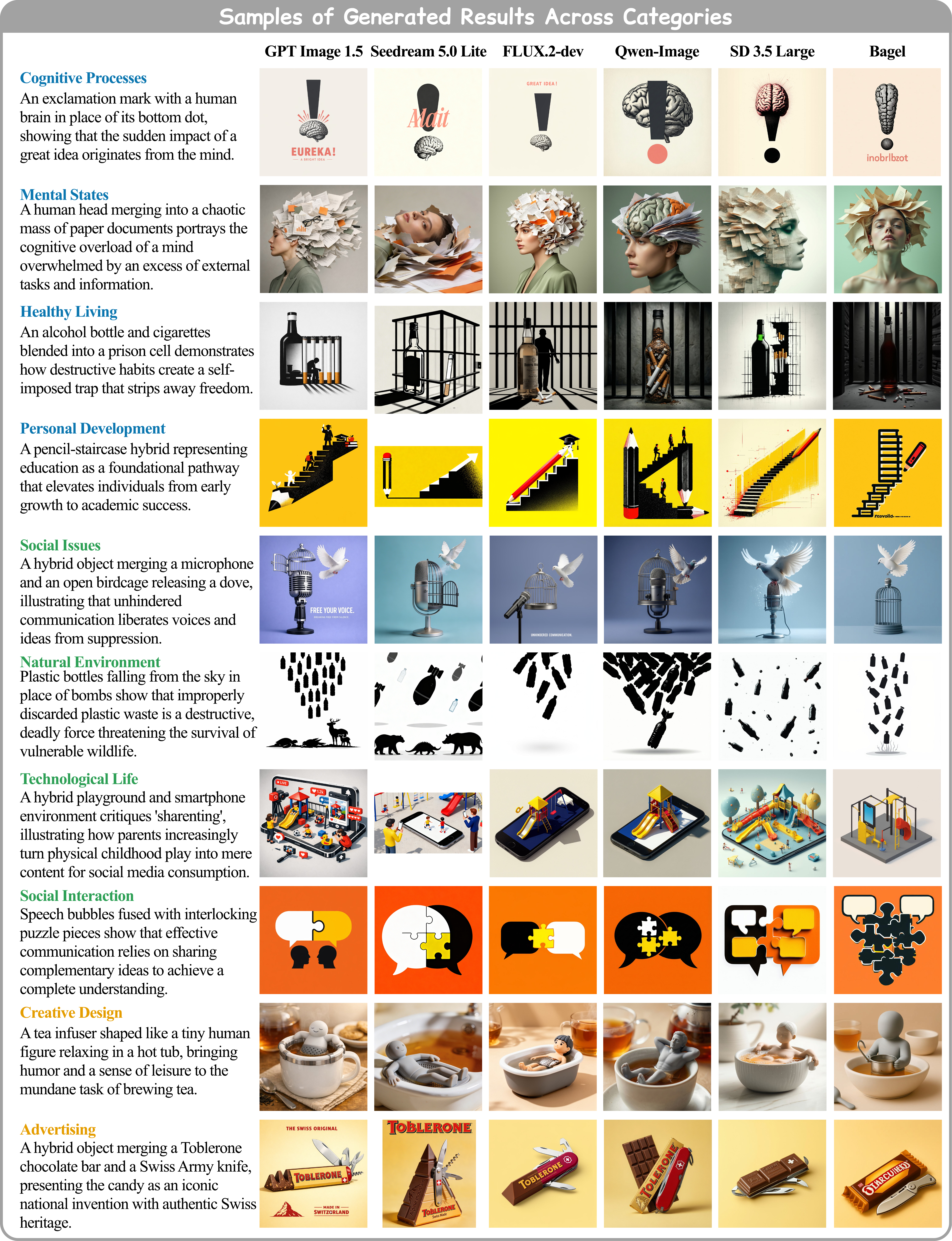}
    \caption{Generated visual metaphors across ten categories from VMetaphor-Bench.}
    \label{fig:samples}
\end{figure}

We present representative samples in Figure~\ref{fig:samples}. For each category, we select one visual metaphor and display the generated images from all evaluated models under the conceptual prompt setting. These examples illustrate the diversity of metaphorical themes in our benchmark and highlight the varying capabilities of different models in generating visual metaphors.
  
\subsection{Limitations}
\label{app:limitations}
We acknowledge several limitations. First, the distribution of samples across thematic categories is uneven, reflecting the natural prevalence of visual metaphors in different topics in real-world creative imagery. Second, our evaluation relies on an MLLM-as-judge framework, which, despite strong alignment with human annotators, inherits the fundamental limitations of automatic visual reasoning on the most subjective judgments. Finally, our prompts and source imagery are predominantly English and from Western creative traditions, leaving multilingual prompts and cross-cultural metaphors as future directions.

\subsection{Broader Impacts}
\label{app:broader_impacts}
This work advances the evaluation of visual metaphor generation, a challenging aspect of text-to-image models that requires coherent cross-domain reasoning and expressive visual composition. By providing a structured benchmark with fine-grained evaluation criteria, it can support the development of more capable and interpretable generative models. Such progress may benefit real-world applications including creative design, education, and advertising, where visual metaphors are widely used to communicate messages effectively and engage audiences.

We also acknowledge potential negative implications. Improved visual metaphor generation could be misused to produce persuasive misleading imagery (e.g., manipulative advertising or political propaganda) that exploits the strong emotional and rhetorical effects of visual analogies. We encourage future work to consider safeguards against such misuse, and our benchmark itself is intended for evaluation rather than direct deployment.

\subsection{Data Licensing and Release}
To respect the rights of original content creators on Pinterest, we do not redistribute the source images. Our public release consists of: (1) image URLs pointing to the original Pinterest pages, (2) all structured annotations (target/source domains, structure types,  meaning, element mappings), (3) descriptive and conceptual prompts, and (4) the 9,594 multiple-choice questions. Users can retrieve images via the URLs for research purposes following Pinterest's terms of service. The annotations and prompts are released under CC-BY-4.0.


\end{document}